\pdfoutput=1
\documentclass[11pt]{article}

\usepackage{acl}

\usepackage{times}
\usepackage{latexsym}

\usepackage[T1]{fontenc}
\usepackage[utf8]{inputenc}

\usepackage{microtype}

\usepackage{inconsolata}

\usepackage{graphicx}
\usepackage{tabularx}
\usepackage{enumerate}
\usepackage{algorithm,algorithmic}
\usepackage{amsmath}
\usepackage{CJKutf8}
\usepackage{multirow}

\title{Eval-GCSC: A New Metric for Evaluating ChatGPT's Performance in Chinese Spelling Correction}

\author{Kunting Li\textsuperscript{1}\thanks{Work was done when Kunting Li was interning at WeChat AI, Tencent Inc, China.}, Yong Hu\textsuperscript{2}, Shaolei Wang\textsuperscript{3}, Hanhan Ma\textsuperscript{3}, Liang He\textsuperscript{1*}, Fandong Meng\textsuperscript{2}, Jie Zhou\textsuperscript{2} \\
  \textsuperscript{1}Department of Electronic Engineering, and Beijing National Research \\Center for Information Science and Technology, Tsinghua University, Beijing 100084, China\\
  \texttt{lkt22@mails.tsinghua.edu.cn, heliang@mail.tsinghua.edu.cn} \\
  \textsuperscript{2}WeChat AI, Tencent Inc, China \textsuperscript{3}Xinjiang University, Urumqi 830017, China \\
  \texttt{\{rightyonghu,fandongmeng,withtomzhou\}@tencent.com \{wangsl0407,danjier\}@stu.xju.edu.cn} 
  \\
}

\begin{document}

\maketitle

\begin{abstract}
ChatGPT has demonstrated impressive performance in various downstream tasks. However, in the Chinese Spelling Correction (CSC) task, we observe a discrepancy: while ChatGPT performs well under human evaluation, it scores poorly according to traditional metrics. We believe this inconsistency arises because the traditional metrics are not well-suited for evaluating generative models. Their overly strict length and phonics constraints may lead to underestimating ChatGPT's correction capabilities. To better evaluate generative models in the CSC task, this paper proposes a new evaluation metric: Eval-GCSC. By incorporating word-level and semantic similarity judgments, it relaxes the stringent length and phonics constraints. Experimental results show that Eval-GCSC closely aligns with human evaluations. Under this metric, ChatGPT's performance is comparable to traditional token-level classification models (TCM), demonstrating its potential as a CSC tool. The source code and scripts can be accessed at \url{https://github.com/ktlKTL/Eval-GCSC}.
\end{abstract}
% ChatGPT已经在各种下游任务中表现出惊艳的效果，但在中文拼写纠错任务中，我们观测到ChatGPT在人工评价下纠错性能良好，但在传统评估指标下却显示其性能很差。我们认为导致这种不一致的原因是CSC任务传统评估指标并不适合评估生成式模型，其过于严格的长度和拼音约束会造成对ChatGPT纠错性能的低估。为了在CSC任务中更好地评估生成模型，本文提出了一种新的CSC评价指标，即Eval-GCSC。通过词级别和语义相似性判断的设计，放松了长度和拼音强约束。实验结果表，Eval-GCSC接近人工评价，且在Eval-GCSC下，ChatGPT性能与传统微调模型可比，是一个有潜力的CSC工具。源代码和脚本可以在链接a上免费获得。

\section{Introduction}
\label{sec:introduction}
\begin{figure}[!ht]
    \centering
    \includegraphics[width=1\linewidth]{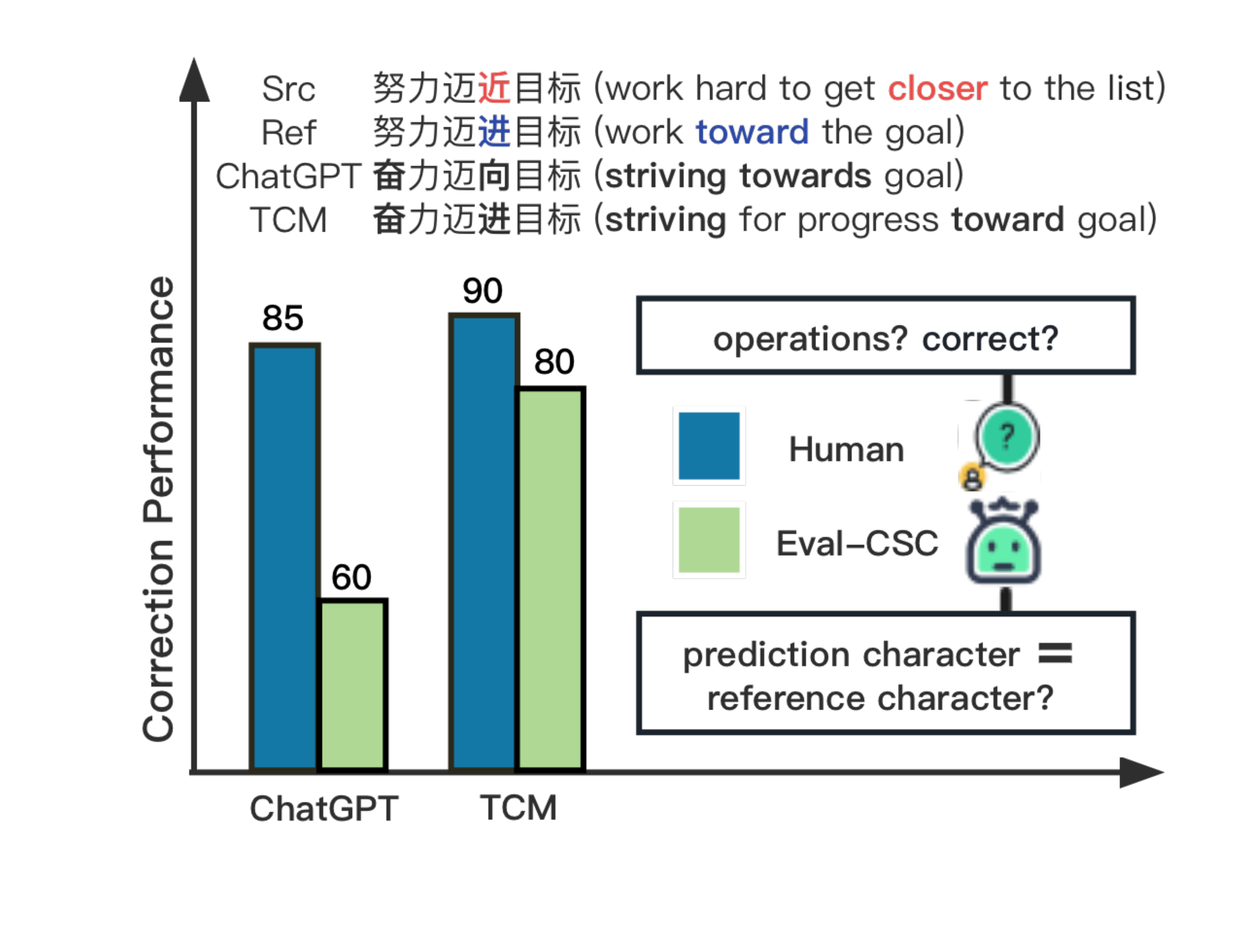}
    \caption{\label{fig:human-eval}The discrepancy in evaluation results under both human evaluation and traditional metrics (we denote by Eval-CSC). "TCM" represents token-level classification model.} 
\end{figure}
% 人工评价和Eval-CSC这两种评估结果的差异。

% 在传统评估指标和人工评价下，ChatGPT和传统微调模型纠错性能的定性对比结果。定性结果显示人工评价下ChatGPT的纠错能力与传统微调模型相差的没有传统指标所显示的那么大。
Chinese Spelling Correction (CSC) is a task that takes a sentence with Chinese spelling errors as input and outputs the corrected sentence. CSC plays a crucial role in various NLP applications \cite{PLOME} and numerous TCM \cite{hong-etal-2019-faspell,PLOME,Phonetic} were researched for CSC task. Nowadays, ChatGPT\footnote{https://chat.openai.com} has demonstrated the impressive capabilities in various NLP tasks \cite{wang2023chatgpt,he2023can}. Increasingly, studies are exploring its capabilities in the field of text error correction \cite{csc_chatgpt,Grammar_chatgpt1,Grammar_chatgpt2}. 
And it has been shown that investigating ChatGPT's CSC abilities can effectively reflect its overall Chinese language processing capabilities \cite{csc_chatgpt}. 
% The main types of Chinese spelling errors are phonetic and visual errors, which impose strong constraints on the output (phonetics, character shape, and length constraints). 
% chatgpt在许多自然语言处理任务上都表现出不俗的能力，其中也有越来越多的工作探究ChatGPT在文本纠错领域的纠错能力。中文拼写纠错（CSC）是一个以中文拼写有错误的句子为输入，输出纠正后的句子的任务。其中中文拼写错误主要分为语音错误和视觉错误，这使任务本身对输出具有了强约束性（拼音、字形和长度约束）。CSC任务在搜索引擎等各种自然语言处理应用中发挥着重要作用，且有工作表明研究ChatGPT的中文拼写能力可以很好地反映其中文处理能力。在ChatGPT出现之前，许多纠错小模型被研究并用于完成CSC任务。目前这些微调模型仍然使用传统的指标进行CSC评估。

Currently, the evaluation metric for CSC (we denote by Eval-CSC) assesses the accuracy of each predicted character by comparing the source and predicted sentences in a left-to-right manner. With the emergence of ChatGPT, many people seem to prefer using it directly for text modification and improvement \cite{Grammar_chatgpt1}. Our tests reveal that ChatGPT's error correction performance under Eval-CSC is poor. However, preliminary manual observations suggest that ChatGPT's error correction is quite effective, comparable to TCM. Figure~\ref{fig:human-eval} qualitatively indicates that the error correction capability of ChatGPT under human evaluation does not deviate as significantly from TCM as suggested by Eval-CSC. Consistent with the analysis in \cite{csc_chatgpt}, we also believe that this inconsistency arises because Eval-CSC is unsuitable for assessing generative error correction models.
% 如今，CSC任务的评估指标（我们称为a）通过从左到右对比原句和预测句判断每个预测字符的正确性，这使得其只能评估等长的纠正操作，不等长的纠正操作则会被忽略。
% 当ChatGPT出现后，似乎许多人喜欢将它直接用于文本修改和完善\cite{abs-2304-01746-tran}。我们测试发现，传统评估指标下ChatGPT的纠错性能很差，但初步人工观测发现chatgpt的纠错效果很好，跟传统模型是可比的。图1展示了两种评价结果的不一致。与论文b的分析一致，我们同样认为导致这种不一致的原因在于传统评价指标并不适合评估生成式纠错模型。

Furthermore, we observe that the Eval-CSC's character-level comparison only allows for the assessment of corrections of equal length, ignoring those of unequal length. This introduces a length constraint. In Eval-CSC, a prediction is only considered correct if the predicted character perfectly matches the reference character. Given that reference characters are often phonetically similar to the source characters, Eval-CSC fail to reasonably evaluate non-phonetic corrections, leading to a phonetic constraint. Moreover, due to the strong reliance on reference sentences, a large number of them must be collected for a reasonable evaluation. However, it is impossible to exhaustively gather all reference sentences.
In contrast, human evaluators typically focus on two aspects: what changes have been made in the predicted sentence compared to the original, and whether these modifications alter the original meaning. Comparing these two evaluation methods, it becomes evident that the strict length and phonetic constraints of Eval-CSC do not necessarily align with the concerns of human evaluators. This discrepancy could lead to an underestimation of ChatGPT's performance, as both unequal length and non-phonetic corrections can be reasonably evaluated under human evaluation.
% 进一步，我们注意到传统评估指标字符级的对比使其只能评估等长的纠正操作，不等长的纠正操作则会被忽略，这是一种长度约束。另外在传统指标中，只有预测字符和参考字符完全一致才算正确，而参考字符往往和原字符是音近的，这导致传统指标无法合理评估非音近操作，这是一种拼音约束。而且由于这种对参考句的强依赖，为了合理评估就需要收集大量参考句，但参考句是不能穷尽的。反观人工评价的过程，人往往只关注两点，一是预测句相比原句改了哪些地方，二是这些修改是否改变原句意思。对比这两种评价方法，可以发现传统指标严格的长度和拼音约束 与人工评价所关心的 没有必然联系，还有可能导致对ChatGPT性能的低估，因为非等长和非音近在人工指标下都可以被合理评估。

% Furthermore, based on the characteristics of Eval-CSC and our exploration of ChatGPT's understanding of length and phonics, we notice that the length and phonetic constraints of Eval-CSC (length constraint refers to maintaining the same length between the predicted and source sentences, while phonetic constraint refers to the predicted characters being phonetically similar to the source characters) are overly strict, differing from the softer constraints of human evaluation. These rigid constraints do not necessarily correlate with the content and correctness concerns of human evaluation. Therefore, using Eval-CSC to assess the performance of ChatGPT in error correction tasks is not reasonable, as it may not only underestimate ChatGPT's performance but also lack a necessary correlation with human evaluation results.

% 进一步，根据传统评价指标的特性和对ChatGPT对长度和拼音的理解能力的探究，我们注意到传统评估指标的长度和拼音约束（长度约束是指预测句和原句长度保持一致、拼音约束是指预测字符与原字符拼音相近）过于严格，而且与人工评估的软约束不同。这两种硬约束与人工评价所关心的操作内容和正确性都没有必然联系，使用其来评估ChatGPT做纠错任务的性能是不合理的，不仅会导致对ChatGPT性能的低估，而且与人工评价的结果也没有必然关联。

To alleviate the aforementioned limitations, we propose Eval-GCSC, a novel \textbf{Eval}uation metric for \textbf{G}enerative error correction models, for the \textbf{CSC}. Building upon Eval-CSC, Eval-GCSC circumvents the constraints and dependency on reference characters by designing word-level and semantic similarity. Mimicking the manual evaluation process, we first design a word-level comparison between the predicted sentence and the source sentence, extracting all operations performed by the predicted sentence relative to the source. Subsequently, we devise a semantic similarity assessment to evaluate the accuracy of each operation. 

We conducted experiments on the widely used benchmark dataset SIGHAN \cite{wu2013chinese,yu2014overview,tseng2015introduction}, manually merging and filtering the SIGHAN13-15 datasets. Experimental results indicate that Eval-GCSC aligns more closely with human evaluation than Eval-CSC. Furthermore, under the assessment of Eval-GCSC, ChatGPT performs well in the CSC task, demonstrating a comparable performance to traditional token-level classification models.
% 实验结果表明，Eval-GCSC比Eval-CSC更贴近人工评估。同时，在Eval-GCSC评估下，chatgpt在csc任务上的表现很好，跟传统模型可比。

Our main contributions are listed as follows:
\begin{enumerate}[(1)]
    \item We propose a new metric Eval-GCSC, for assessing the error correction performance of generative models in CSC task, which shows strong correlation with human evaluations.
    \item Under Eval-GCSC, ChatGPT exhibits comparable correction performance to traditional token-level classification models, indicating its potential as a CSC tool.
    \item We analyze the strengths and weaknesses of ChatGPT in CSC tasks. Specifically, we found that although ChatGPT occasionally introduces errors in the corrective output for the CSC task, it generally produces semantically more appropriate corrections than traditional classification models. These findings provide valuable insights for future research.
\end{enumerate}

% 1> 我们提出了用于评估生成模型在CSC任务中的纠错性能的新指标Eval-GCSC，其与人工评估具有良好的相关性。
% 2> 在新指标下，我们发现ChatGPT与传统纠错微调小模型具有可比性，是一个有潜力的CSC工具。
% 3> 我们根据新指标评估结果分析发现：

% <ChatGPT是有潜力的CSC工具> 为了分析Eval-GCSC评估下ChatGPT在CSC任务中与微调小模型的性能接近的原因，我们对同一评价指标Eval-GCSC(GPT-4)下ChatGPT和微调模型的纠错结果进行了详细的观察对比。为更好地对比ChatGPT和微调模型，我们选取所有两种模型对原句相同位置进行修改的纠正操作，并分下面三种情况分析：
% 1》ChatGPT纠正正确，微调模型纠正错误：我们发现60%的样例中chatgpt的纠正操作比小模型语义更合适，另外20%的样例的参考句语义模糊，但chatgpt改得更合理；其余少部分样例则体现chatgpt对连续两个错别字可以更好地修改，且chatgpt对知识性的错误能更好地纠错。
% 2〉ChatGPT纠正错误，微调模型纠正正确：61%的情况下chatgpt倾向改变句子原意，有时会顺着错误词的第一个字造词，有时则按照自己的理解修改语义；不改变语义情况下，chatgpt存在纠正不正确的情况（占26%）；还有少部分样例中，chatgpt会由于增加或删除字符导致语义不合适。
% 3》ChatGPT和微调模型均纠正错误：这些样例主要是原句存在如下情况：知识型错误、不常用的词语、连着两个错别字造成词语意思不清晰、原句语义不清晰、错别字位置可能存在多种合理的修改。

% 总之，ChatGPT极强的语言处理能力使其做CSC任务具有一定的优势，但仍存在不足的原因可能是其优化策略更偏向于生成流畅、自然的文本，而不是严格的拼写准确性，另外其单向生成的方式，无法在生成文本后进行回溯和修改，这也可能限制了其纠错能力。

\section{Related works}

\textbf{Basic Competence of ChatGPT}
The study \cite{csc_chatgpt} found that while ChatGPT knows the phonics of Chinese characters in the CSC task, it does not understand how to pronounce them, making it challenging to correct pronunciation errors. In the grammar error correction task, \cite{Grammar_chatgpt2} evaluated ChatGPT and found that it tends to overcorrect, which may stem from its diverse generative capabilities as a large language model. \cite{Grammar_chatgpt1} also indicated that ChatGPT freely corrects errors, resulting in highly fluent revised sentences. This could be due to its tendency to overcorrect and its disregard for the principle of minimal editing. These conclusions are consistent with the observations of our experiments.
% 工作b发现ChatGPT在CSC任务中知道汉字的拼音，但不能理解如何发音，因此很难纠正语音错误；工作b在语法纠错GEC任务下对ChatGPT进行评估，发现ChatGPT会自由地纠正错误使更正的句子非常流畅产生更多的过度纠正，这可能来自于作为一个大型语言模型的多样化的生成能力；研究b同样表明ChatGPT会自由地纠正错误使更正的句子非常流畅，这可能是由于它的过度更正倾向和不遵守最小编辑原则。这些结论与我们实验的观察一致。

% model
\textbf{Error Correction Models} Current error correction models primarily fall into two categories. The first category is token-level classification models \cite{hong-etal-2019-faspell,PLOME,Phonetic}, which are based on BERT \cite{devlin-etal-2019-bert} and model CSC as a character-level sequence labeling task. The second category is generative models \cite{lewis-etal-2020-bart}, which correct text errors (mainly grammatical) through generation. In this paper, we select the token-level classification model, which is more suitable for CSC tasks, as our baseline model.
% 目前纠错模型主要分为两类，一类是字符级分类模型，这类模型以BERT为基础，将CSC建模为字级别序列标注任务；另一类是生成模型，以生成的方式进行文本纠错（主要是语法纠错）。本论文中选择更适用于CSC任务的字符级分类模型作为基线模型。

% Metric
\textbf{Evaluation Metrics}
The traditional evaluation metrics for CSC task, although implemented differently across various open-source projects \cite{PLOME,xu-etal-2021-read,cheng-etal-2020-spellgcn,wang-etal-2019-confusionset}, share a common underlying principle. The common practice for CSC evaluation is to compare the system output with the corresponding reference and then use the calculated precision, recall, and F1 scores as evaluation metrics. But computation of these metrics overly relies on reference sentences. To solve this problem, referring to the ideas in \cite{BERTScore}, we adopt semantic similarity in designing CSC metrics.
% CSC任务的传统评价指标，在各个开源项目中虽然写法不同，但基本思想是相同的。CSC评价的通常做法是判断系统输出与相应参考是否一致，进而将计算出的精确率、召回和F1分数作为评价指标。
% To reduce the cost of collecting reference sentences, \textbf{many studies} have designed evaluation metrics based on semantic similarity. 
% Similar to the CSC task, machine translation tasks also calculate metrics by comparing predicted sentences with reference sentences.
% 这些传统指标的计算均过度依赖参考句。与CSC任务类似，机器翻译任务也通过对比预测句和参考句计算指标。为减小收集参考句的成本，许多工作设计基于语义相似度的评估指标。参考这个思想，我们在设计CSC指标时也采用了语义相似度。

Furthermore, \cite{csc_chatgpt} argues that the widely used traditional metrics in Chinese text correction tasks cannot accurately reflect the correction capabilities of LLMs. Designing and developing new evaluation metrics for LLMs is an essential and valuable future direction. In this work, we designed new CSC evaluation metrics and tested the correction performance of models under these new metrics, proving the potential of ChatGPT to perform CSC tasks.
% 然而，论文a提出中文文本校正任务中广泛使用的传统自动评价指标并不能真实客观地反映LLMs的校正能力，针对LLMs设计和开发新的评价指标是一个重要且有价值的未来方向。在这项工作中，我们设计了新的CSC评估指标，并通过实验测试了新指标下各个模型的纠错性能。

\section{Motivation}

\subsection{ChatGPT Basic Competence Tests}
\label{sec:Basic Competence}
% 不同词语长度下，ChatGPT在长度和拼音测试中的正确率。
% chatgpt对长度概念不清晰 
% \textbf{Basic Competence Tests for ChatGPT} 
% In the calculation of Eval-CSC, it is crucial to ensure that the length of the predicted sentence matches the original and that the characters predicted are phonetically similar to those in the source sentence. To have the model's output satisfy these length and phonics constraints, the model needs to have a clear understanding of both. 
% ChatGPT is a generative large-scale model, and its method of predicting the next token makes it challenging to intervene in the generation results. 
To investigate whether ChatGPT can effectively understand the concepts of length and phonics\footnote{Phonetic errors dominate in CSC task \cite{zhang-etal-2020-spelling,PLOME}, so we focused our testing solely on phonics, disregarding character shape.} required for CSC tasks, we designed two types of tests for both length and phonics: generation and discrimination. The content is as follows, and specific procedures can be found in the Appendix~\ref{app:length and pho eval}.

\textbf{Length Test} The generation tests require ChatGPT to output the length of Chinese characters, and the discrimination tests require identifying whether two words have the same length.

\textbf{Phonics Test} The generation tests require ChatGPT to output the characters with similar phonics, and the discrimination tests require identifying whether two words have identical phonics. 

% Specifically, the length and phonics generation tests require ChatGPT to output the length of Chinese characters and characters with similar phonics, respectively. The length and phonics discrimination tests require to identify whether two words have the same length or identical phonics. The specific procedures and design of the prompts can be found in the Appendix~\ref{app:length and pho eval}.

% 在计算传统指标时，保证预测句长度与原句一致、预测字符与原句字符拼音相近是关键。若想让模型的输出满足长度和拼音约束，需要模型对两者有清晰的概念。
% ChatGPT属于一种生成式大模型，其预测下一词的生成方式导致人们很难干预生成结果。为探究ChatGPT能否很好地理解CSC任务所需要的长度和拼音的概念，我们对长度和拼音分别设计了生成和辨别这两种测试。具体来说，长度、拼音生成测试分别需要chatgpt输出汉字的长度、拼音相近的汉字，长度、拼音辨别测试则分别需要辨别两个词长度是否一致、两个词拼音是否相同。具体prompt的设计见表格。
% 语音错误在CSC任务中占主导地位，故我们只对拼音进行了测试而忽略了字形。

\begin{figure}[htbp]
    \centering
    \includegraphics[width=1\linewidth]{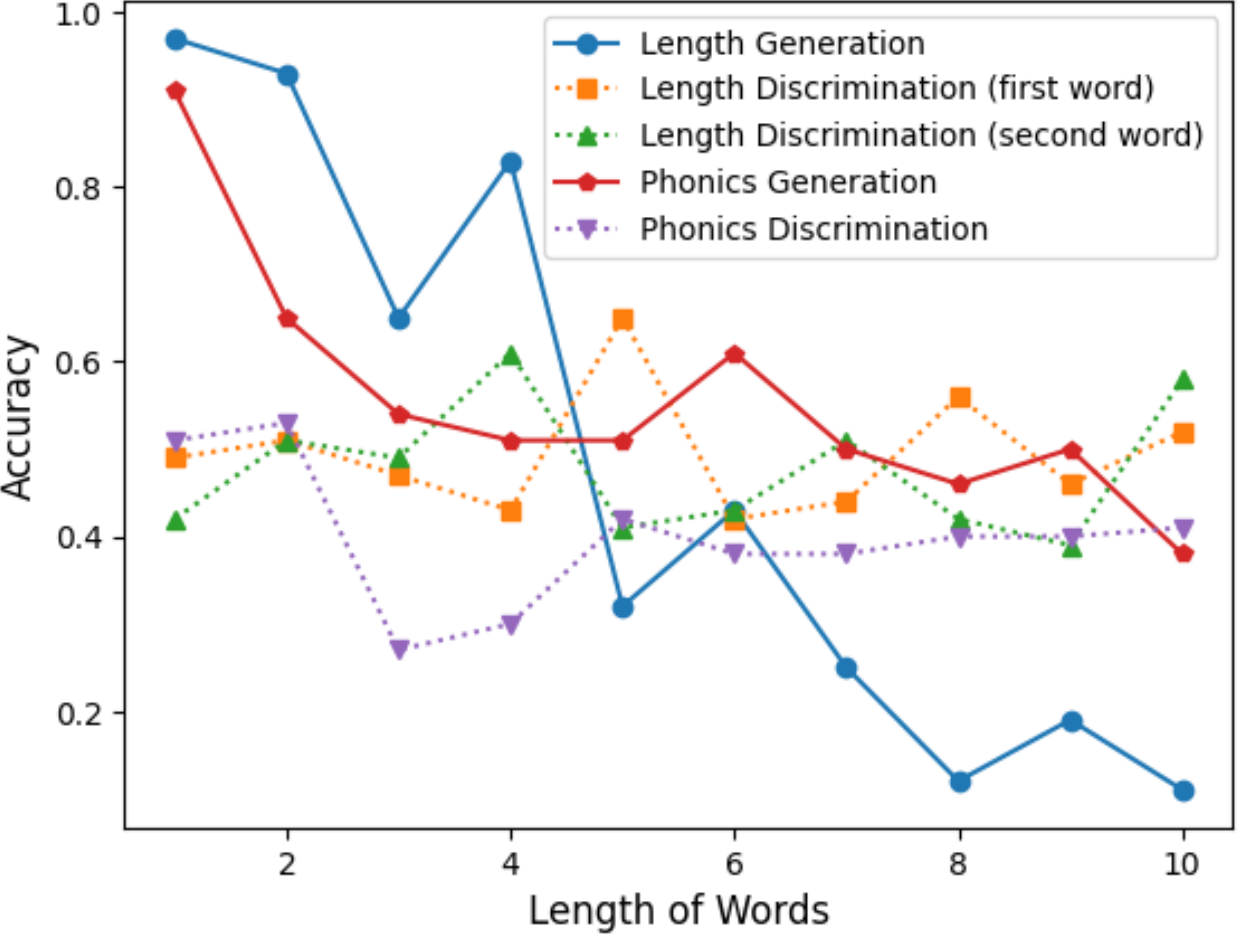}
    \caption{\label{fig:length-pho-eval}The accuracy of ChatGPT in tests of length and phonics under different word lengths.}
\end{figure}

Figure~\ref{fig:length-pho-eval} shows that in the generation tests, the accuracy is higher when the word length is shorter, but performance declines significantly as length increases. In the discrimination tests, the overall accuracy is not high. This indicates that ChatGPT struggles to comprehend the concepts of length and phonics effectively. Possible reasons include that language models typically use tokenizers to parse text, resulting in mediocre performance in character computation. Additionally, phonics without tonal marks can easily be confused with English, and the model may struggle to map Chinese characters to their corresponding phonics, which could account for the poor understanding of phonics.
% 从结果来看，生成测试中词语长度较短时，正确率较高，随着长度增加，性能下降明显；辨别测试中，整体上正确率都不高。这说明ChatGPT不能很好地理解长度和拼音的概念。可能的原因是LLM一般使用分词器（tokenizer）解析文本，这使它们往往在计算字符方面表现一般般；另外不带音调的拼音和英文容易混淆，模型不能很好地将汉字和拼音对应起来，这可能是导致拼音理解能力差的原因。

\subsection{Human Evaluation for CSC}
Analyzing the process of human evaluation provides significant insights for designing assessment metrics. Figure~\ref{fig:human-eval} illustrates the two main steps involved in the human evaluation of error correction results: firstly, observing the modifications made in the predicted sentence compared to the original, and whether the typos in the source sentence have been corrected; secondly, determining whether the modifications in the predicted sentence have altered the original meaning. Intuitively, these two aspects of concern in human evaluation are also of interest in applying CSC.
% 分析人工评价的过程对于设计评估指标有重要启发作用。图1展示了人工评估纠错结果时主要进行的两个评估步骤：首先是观察预测句相比原句有哪些修改操作、原句中的错别字是否被修改；其次判断预测句的修改操作是否改变了原句的意思。直觉上，人工评价所关心的这两点，也是CSC任务在应用中所关心的。

\begin{figure*}[htbp]
\begin{center}
\includegraphics[width=1\linewidth]{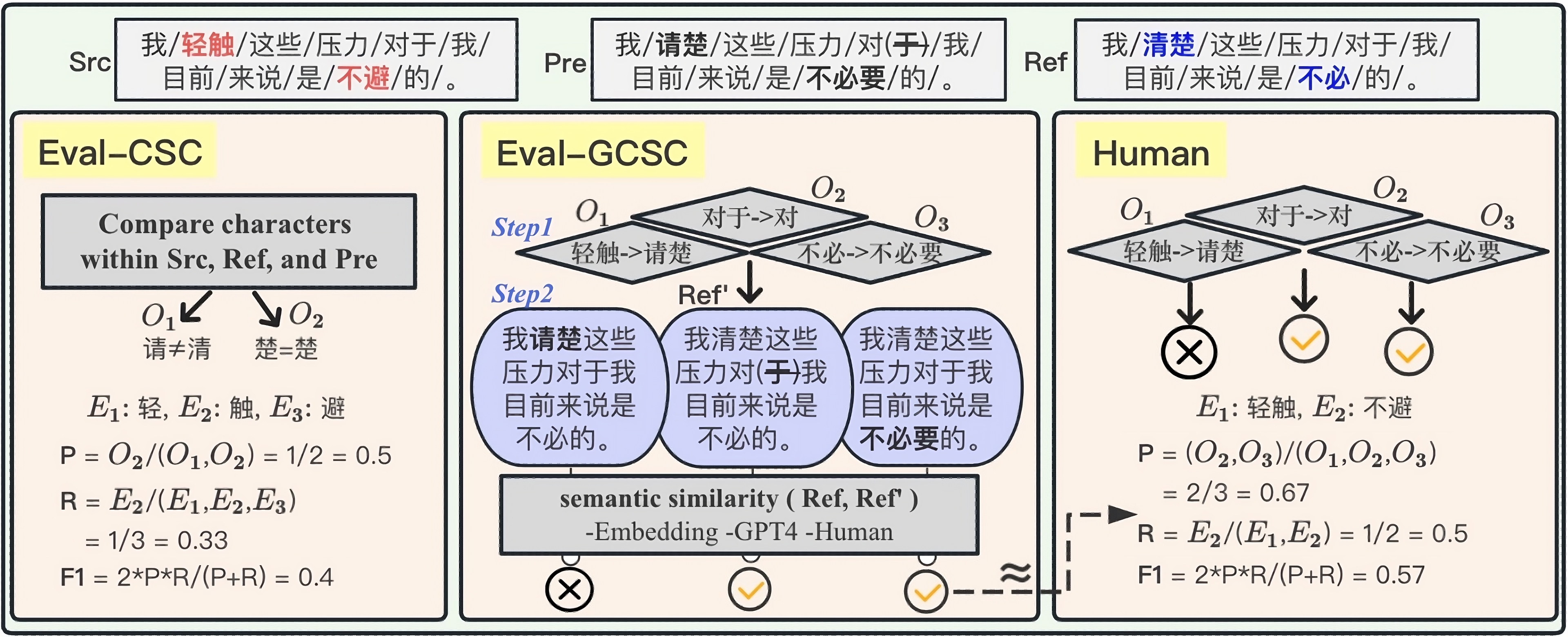}
\caption{\label{fig:old and new metric}Overview of the calculation process for traditional evaluation metrics (Eval-CSC) and our proposed metric (Eval-GCSC), with $Src$, $Ref$, $Ref'$ and $Pre$ representing the source sentence, the reference sentence, the processed reference sentence, and the predicted sentence, respectively. Our method consists of two core modules: 1) extracting edit operations from the corresponding sentence pairs; 2) assessing the correctness of each operation based on semantic similarity, which can be obtained from sentence embeddings, GPT4, or human evaluation. We mark the \textcolor{red}{\textbf{wrong}}/\textcolor{blue}{\textbf{correct}} characters.}
\end{center}
\end{figure*}

\subsection{Revisiting Eval-CSC}
\label{sec:old metric}
\textbf{Length Constraint} In the task of CSC, previous studies have used sentence-level and character-level precision, recall, and F1 scores as Eval-CSC. These metrics require a character-level comparison between the source and predicted sentences to identify all correction operations, necessitating that the source, reference, and predicted sentences be of the same length. This may result in ignoring operations that lead to a length discrepancy with the source sentence. However, we argue that this strict length constraint is not only overly rigorous and inconsistent with practical applications, but also unrelated to human evaluation's two aspects of concern (the operations and their correctness). The impact of modification operations on the source sentence length is hardly considered in human evaluation.

Further, we used the ChERRANT \cite{zhang-etal-2022-mucgec} to count all modification operations in the predicted sentences compared to the source sentences under zero-shot and few-shot conditions (prompt seen in Table~\ref{tab:Zero-shot and few-shot prompt}) for ChatGPT. The results are shown in Table~\ref{tab:old metric operation}. Approximately 30\% of all operations involve unequal length operations (including additions, deletions, and unequal length replacements), due to ChatGPT's limited understanding of length concept as mentioned in Section~\ref{sec:Basic Competence}. Among these unequal length operations, about 93\% were judged correct by human evaluators. This suggests that Eval-CSC, due to its length constraint, overlooks correct unequal length operations, leading to an underestimation of ChatGPT's error correction performance. Conversely, while ChatGPT may perform poorly under traditional metrics due to unequal length operations, it does not necessarily imply poor performance under human evaluation, as unequal length operations can also be correct.

\textbf{Phonics Constraint} In CSC tasks, reference characters and source characters are mostly phonetically similar, and Eval-CSC requires the predicted character to be identical to the reference character to be considered correct. This makes error correction performance heavily dependent on the reference sentence, increasing the cost of collecting multi-reference data for a more comprehensive evaluation. Similar to our previous analysis, we believe this strong phonetic constraint is unrelated to the concern in human evaluation (the operations and their correctness). Whether a modification operation is phonetically similar to the source word is hardly considered in human evaluation. 

Furthermore, ChatGPT's limited understanding of phonics may make non-phonetic modifications (statistics seen in Table~\ref{tab:old metric operation}). The results show that 20\% of equal length replacements involve non-phonetically similar operations, but 40\% of these operations are deemed reasonable under human evaluation. This suggests that Eval-CSC, due to its phonetic constraint, misjudges correct non-phonetically similar operations, leading to an underestimation of ChatGPT's error correction performance. Conversely, while ChatGPT may perform poorly under traditional metrics due to non-phonetically similar operations, it does not necessarily imply poor performance under human evaluation, as non-phonetically similar operations can also be correct.

% In the CSC task, the reference characters and source characters are mostly phonetically similar. When calculating Eval-CSC, it is required that the predicted character and the reference character must be exactly the same to be considered correct. ChatGPT may not understand the concept of phonics well, and its modifications to the source sentence may not be phonetically similar to the source characters (statistics shown in Table~\ref{tab:old metric operation}). The results show that in equal-length replacements, non-phonetically similar operations account for 20\%. These non-phonetically similar predicted characters are judged as errors because they do not match the reference characters, even if their semantics might be appropriate. This is another potential underestimation of ChatGPT's error correction performance. Moreover, the error correction performance heavily depends on the reference sentence, and the predicted character must be exactly the same as the target character to be considered correct. This means that if we want to comprehensively evaluate the error correction performance, it will inevitably increase the cost of collecting multi-reference data.
% 《拼音约束》在CSC任务中，参考字符和原字符大多是拼音相近的，计算Eval-CSC时要求预测字符和参考字符完全一致才算正确。这使纠错性能对参考句子的依赖性很大，如果希望更全面评估纠错性能，必定会增加收集multi-reference数据的成本。与前面分析类似，我们认为这种拼音的强约束与人工评估关心的两点(操作以及操作的正确性)也没有必然联系，修改操作是否与原词拼音相近在人工评价中几乎不会被关注。另外ChatGPT由于无法很好的理解拼音的概念，对原句的修改可能与原字符非音近（统计结果见表a），统计结果显示在等长替换中非音近的操作占20%，但非音近操作中b%的操作在人工评价下是合理的。这说明Eval-CSC由于拼音约束会误判正确的非音近操作，这造成对ChatGPT纠错性能的低估；另一个角度，ChatGPT虽然在传统指标下可能由于非音近操作导致性能不佳，但不代表在人工评估下性能也会不佳，因为非音进操作也可能在人工评估下是正确的。

\begin{table}[t!]
    \centering
    \resizebox{\linewidth}{!}{
     \begin{tabular}{c|ccccc|c}
     \hline
    ChatGPT               & $R$   & $M$   & $S_{ne}$ &  $C_{ne}$ & $S_{enp}$ & $C_{enp}$\\ \hline
    Few-Shot  & 11.47 & 8.46 & 4.56  & 93.12  & 19.69  & 41.20        \\
    Zero-Shot & 18.07 & 14.02 & 4.96 & 91.88  & 21.92  & 39.15    \\ \hline
    \end{tabular}
    }
\caption{\label{tab:old metric operation}The proportion of ChatGPT's operations in CSC. Detailed experimental settings can be found in Section~\ref{sec:exp setup}. $M$ and $R$, respectively, denote insertion and deletion operations. $S_{ne}$ represents non-equal length substitution operations, while $S_{enp}$ refers to equal length substitution operations that are not phonetically similar. $C_{ne}$ and $C_{enp}$ respectively represent the proportion of operations deemed reasonable in non-equal length operations and non-homophonic equal length operations, as assessed by human.}
\end{table}
% BERT                & 2  & 4   & 1   & 609    & 2750  & 2368          \\
% SoftMaskedBERT      & 0  & 619 & 0   & 18     & 2700  & 2405          \\
% PLOME               & 1  & 2   & 1   & 785    & 2498  & 2383  \\ \hline

In summary, we believe using Eval-CSC to assess the error correction performance of generative models like ChatGPT is inappropriate. This approach not only tends to underestimate the model's performance, but its stringent length and phonetic constraints also differ from the softer constraints of human evaluation. The aspects that human evaluators focus on, such as "what is the operation" and "is the operation correct", do not necessarily correlate with these strict constraints. Whether the correction is phonetically similar to the source word or maintains the same length is generally not a concern in human evaluation. Therefore, poor performance of ChatGPT under Eval-CSC due to length or phonetics does not necessarily imply poor performance under human evaluation. 
% Compared to these rigid constraints, the softer constraints of human evaluation provide a more accurate reflection of the model's effectiveness.
% Furthermore, in practical applications, minor length variations and non-phonetically similar modifications are acceptable as long as they maintain semantic consistency. For instance, when a search engine corrects a user's spelling errors, as long as the meaning is not affected, users will not be bothered by inconsistencies in the length or phonetics of the sentence before and after the correction.

%  长度约束 -》 大量非等长操作 -〉低估。 拼音约束 -》 大量非音近情况 -〉低估。

% 总之，我们认为Eval-CSC来评估ChatGPT这种生成模型的纠错性能是不合理的，因为其不仅会造成对性能的低估，而且其过于严格的长度和拼音强约束与人工评价是有差别的，相比长度和拼音的硬约束，人工评价这种软约束更能真实反映模型效果。人工评估所关心的“操作是什么”以及“操作的正确性”与这两个强约束没有必然联系，修改操作是否与原词拼音相近、长度一致在人工评价中都几乎不会被关注。ChatGPT在Eval-CSC下可能由于长度或拼音导致的性能不佳，并不代表人工评价下性能会不佳。此外在实际应用中，保持语义一致的前提下，直觉上，适当的长度变化和非音近的修改是可以接受的，如在搜索引擎中纠正用户输入的拼写错误时，只要不影响语义，用户不会纠结于 原句长度或拼音 在改动前后的不一致。

\section{Eval-GCSC}
\label{sec:new metric}
We design a new evaluation metric, called Eval-GCSC, which is inspired by the process of human evaluation. This metric involves two steps, as illustrated in Figure~\ref{fig:old and new metric}. Firstly, all word-level operations in the predicted sentence are preserved through a word-level design. Secondly, a semantic similarity assessment is performed to estimate the reasonableness of each correction operation.
% We design an evaluation metric for generative spelling correction models, named Eval-GCSC, inspired by the process of human evaluation. As shown in Figure~\ref{fig:old and new metric}, Eval-GCSC involves two steps: initially, all word-level operations in the predicted sentence are preserved through a word-level design, followed by a reasonableness estimation for each correction operation via semantic similarity assessment.
% , thereby avoiding strong dependence on the reference sentence.
% 仿照人工评价的过程，我们设计了适用于生成式纠错模型的评价指标Eval-GCSC。如图3所示，Eval-GCSC具体分为两个步骤：首先通过词级别的设计保留预测句所有的词级别操作，之后通过语义相似性评估对每个纠正操作进行合理性估计。

\subsection{Word-level Operations Extraction}
\label{sec:Operations Extraction}
% 词级别的设计方法
As previously mentioned, ChatGPT often generates predictions with many non-equal length operations, which complicates the character-level extraction of all operations in the prediction sentence during the calculation of Eval-CSC. This is due to the difficulty in aligning the source and predicted characters. Therefore, we have designed a word-level correction operation extraction method specifically for ChatGPT. In this word-level extraction, words are treated as basic units. 
% generative large models like

Firstly, the reference sentence is segmented using LAC \cite{jiao2018LAC}, and the source sentence is segmented according to the segmentation results of the reference sentence. Subsequently, the Cherrant \cite{zhang-etal-2022-mucgec} is used to obtain the character-level operation position information of the prediction sentence and the character of the operation. The segmentation position of the prediction sentence is obtained as shown in Algorithm~\ref{alg:Framwork word}. Finally, based on the segmentation position of the source and predicted sentence, the corresponding relationship between the source word, reference word, and predicted word is established.
% 正如前文所述，ChatGPT预测句中存在大量的非等长操作，这导致在计算Eval-CSC时字符级地提取预测句所有操作的过程中无法很好地对齐原句字符和预测字符，故我们针对ChatGPT这种生成式大模型设计了一种词级别的纠正操作提取方法。词级别的提取是将词看作基本单元.(利用Cherrant工具得到每个操作对原词的操作信息。具体的计算流程是)首先使用LAC将参考句分词，原句按参考句的分词结果进行分词切割；之后，使用Cherrant得到预测句相比于原句的字符级操作位置信息和操作前后的字符信息，预测句的分词位置利用操作造成的长度变化对原句分词位置修正得到；根据原词和预测句分词位置，得到原词、参考词和预测词的对应关系。公式化表述如下：

The design at the word level not only addresses the issue of poor alignment between source and predicted characters, but also is not constrained by length, preserving every modification made by ChatGPT to the source sentence. This avoids the evaluation bias by ignoring operations that do not maintain equal length.
% 词级别的设计，既解决了无法很好地对齐原句字符和预测字符的问题，也不受长度约束限制，保留了ChatGPT对原句的每一处改动，避免了忽略非等长操作带来的评估偏差。

\renewcommand{\algorithmicrequire}{ \textbf{Input:}} 
\renewcommand{\algorithmicensure}{ \textbf{Output:}} 
\begin{algorithm}[htbp]
\caption{Framework of Operations Extraction.}
\label{alg:Framwork word}
\begin{algorithmic}[1] %这个1 表示每一行都显示数字
\REQUIRE ~~\\ %算法的输入参数：Input
    segmentation position of reference sentence, $I_r = [x_0,x_1,\cdots,x_n]$; The modification start position and termination position of each operation in the source sentence, $P_{start}, P_{end}$; The words processed by each operation, $V$;
\ENSURE ~~\\ %算法的输出：Output
    segmentation position of predict sentence, $I_p$;
\IF{$P_{start} \leq x_i \leq P_{end} (i=0,1,\cdots,n)$}
    \STATE delete $x_i$ in $I_r$; update $I_r$;
    \ENDIF
    \STATE Initialize $I_p=[x'_0,x'_1,\cdots,x'_n]$ as $I_r$;
    \STATE Find $[x_s,x_e] \supseteq [P_{start},P_{end}]$;
\FOR{$j$ in range$(e,n)$}{
    \STATE $x'_j = x'_j + [length(V) - (P_{end} - P_{start})]$
    } %算法的返回值
    \ENDFOR
\RETURN $I_p$.
\STATE Obtain the words in the reference sentence from $x_i$ to $x_{i+1}$, and corresponding words in the predicted sentence from $x'_i$ to $x'_{i+1}$.
\end{algorithmic}
\end{algorithm}

\subsection{Semantic similarity evaluation}
\label{sec:Semantic similarity}
Upon obtaining word-level operation information, we designed a method to traverse all predicted words, integrating each one back into its corresponding position in the reference sentence. This process yields a reassembled reference sentence. We then compare the semantic similarity between the source and reassembled reference sentences. If the semantics are close, the operation is deemed reasonable. We employ three methods to judge the semantic similarity between the reference sentences before and after reassembly: 

\begin{enumerate}[(1)]
    \item Cosine similarity of sentence embedding: The sentence vector is obtained from either Sentence-Transformer \cite{reimers-gurevych-2020-making} or Text2Vec \cite{Text2vec}. The cosine similarity threshold is set as a hyperparameter. If the similarity equals or exceeds this threshold, the semantics are considered similar and the modification operation is deemed reasonable.
    \item GPT-4 evaluation: We use prompt (see Appendix~\ref{app:prompt gpt4}) to directly ask GPT-4 to judge whether the semantics of the two sentences are close. The temperature is set to 0.
    \item Human evaluation: Each operation requires human annotation to determine its correctness. Detailed can be found in Appendix~\ref{app:human eval}. The human evaluation mentioned here is identical to the one mentioned in Section~\ref{sec:introduction}.
\end{enumerate}

% 得到词级别的操作信息后，我们设计遍历所有的预测词，将每个预测词分别拼回参考句中对应的参考词的位置得到拼接后的参考句，然后对比原参考句和拼接后的参考句的语义是否接近，若语义接近说明该操作合理，反之则不合理。我们采用三种方式判断拼接前后参考句的语义相似性，分别是句向量余弦相似度、GPT4评估和人工评价。句向量余弦相似度中，每一句的句向量由sentence-transformer或text2vec得到，余弦相似度阈值设为超参数，当相似性大于等于此阈值即认为语义相似并且修改操作合理。GPT4评估中，则使用提示词（见表1）直接让GPT4判断两句话的语义是否接近。而在人工评价中，需要人工标记每个操作是否正确，具体信息见附录a。

Compared to the strict constraint that requires the predicted character to match the target character, this softer constraint based on semantic judgment reduces the dependency on the reference sentence. This not only lowers the cost of manually collecting numerous reference sentences, but also better evaluates the performance of large generative models in error correction tasks.
% 相比必须要求预测字符与目标字符一致的强约束，这种语意判断的软约束方式，使得预测结果不强依赖于目标句，降低了人工收集许多目标句的成本，同时更好地评估生成式大模型在纠错任务下的性能。

\subsection{Calculation of Eval-GCSC}
\label{sec:Calculation of Eval-GCSC}
% 参考传统的指标设计，我们也计算 词级别 和 句子级别的指标
Based on the design of the above two parts, The specific algorithm process is shown in Algorithm~\ref{alg:Framwork}. Referring to the traditional metrics design, we also calculate sentence-level and character-level precision, recall, and F1 scores (See Appendix~\ref{app:Metric design} for design details). 
% 具体的算法流程如下:
\renewcommand{\algorithmicrequire}{ \textbf{Input:}} 
\renewcommand{\algorithmicensure}{ \textbf{Output:}} 
\begin{algorithm}[ht!]
\caption{Framework of Eval-GCSC.}
\label{alg:Framwork}
\begin{algorithmic}[1] %这个1 表示每一行都显示数字
\REQUIRE ~~\\ %算法的输入参数：Input
    source sentence, $S$; reference sentence, $R$; predict sentence, $P$; segmentation position of reference sentence, $I_r$; segmentation position of predict sentence, $I_p$;
    A set of character-level operations of the predicted sentence compared to the source sentence, $Ops$;
    The modification start position and termination position of each operation in the source sentence, $P_{start}, P_{end}$; The words processed by each operation, $V$;
\ENSURE ~~\\ %算法的输出：Output
    Whether the current operation is correct, $C$;
    \STATE Use $I_r$ to segment $S$ into words; Record the position of all the error words in $S$.
    \STATE According to $I_r$ and $Ops$, update $I_r$ using the method in Algorithm~\ref{alg:Framwork word};
\FOR{$Op$ in $Ops$}{
        % \STATE $I_p$ is assigned the value of $I_r$. Subsequently, $I_p$ is updated based on the modified position $P_{start}$, $P_{end}$ and the length of $V$; 
        % \STATE Based on $P_{start}$ and $P_{end}$, the corresponding segmentation positions $(S_r,E_r)$ in $I_r$ and $(S_p,E_p)$ in $I_p$ are obtained;
        \STATE $I_p$ is obtained according to Algorithm~\ref{alg:Framwork word}. The corresponding positions $(S_r,E_r)$ in $I_r$ and $(S_p,E_p)$ in $I_p$ are also obtained;
        \STATE Reintegrate the predicted word into $R$, yielding the processed reference sentence $R_{pro}$. $R_{pro} = R[:S_r] + P[S_p:E_p] + R[E_r:]$;
        \STATE Evaluate the semantic consistency between $R$ and $R_{pro}$. If they align semantically, it indicates the $Op$ is correctly executed $(C= True)$, otherwise $C=False$;
        \RETURN $C$;} %算法的返回值
    \ENDFOR
\end{algorithmic}
\end{algorithm}
% 详细细节见附录a

Unlike traditional evaluation metrics, we assess the correctness of error correction by predicting the semantic similarity between the reference sentence and the concatenated reference sentence. This means that the predicted word does not necessarily have to match the target word. The source word may be correct, and the modified word is also correct, resulting in the numerator of precision and recall not being equal. 
% 不同于传统评估指标，由于我们通过目标句和拼接后目标句的语义相似性来判断纠错的正确性，预测词不一定与目标词一致，也就是可能存在原词本身无错同时修改的词也正确的情况，这使得精确率和召回率的分子不相等。

\section{Experiments}

\subsection{Experimental Setup} \label{sec:exp setup}

\textbf{Dataset} Experiments were conducted on the widely used benchmark dataset SIGHAN13$\sim$15 \cite{wu2013chinese,yu2014overview,tseng2015introduction}. 
To mitigate the bias in performance across different datasets, we have merged the SIGHAN13$\sim$15 datasets. Subsequently, due to numerous semantically incoherent reference sentences in the SIGHAN dataset, we manually filtered and selected a total of 5071 train samples and 2898 test samples. This avoids the problem of inaccurate estimation of ChatGPT's Chinese spelling correction ability to some extent.
% 为了避免不同数据集下性能存在的偏差，我将SIGHAN13$\sim$15合并。之后由于原始SIGHAN数据集中存在大量语义不通顺的真值句子，我们从中人工过滤，筛选出共2898个样本，以便在一定程度上避免对ChatGPT中文拼写纠错能力的错误估计。

\textbf{Models} To compare the performance of ChatGPT, we selected three token-level classification CSC models as our baselines. Besides, we evaluated the capabilities of ChatGPT in CSC from both zero-shot and few-shot perspectives.
\begin{enumerate}[(1)]
    \item Baseline: BERT \cite{devlin-etal-2019-bert} is a bidirectional language model, employs a classifier to select the corrected character output from the vocabulary for each character; SoftMaskedBERT \cite{zhang-etal-2020-spelling} (SMBERT) addresses the limitation of BERT in detecting errors at each position by incorporating an error detection network and a BERT-based error correction network; PLOME \cite{PLOME} utilizes phonics and strokes as inputs for both the pre-training and fine-tuning, designing character prediction tasks and phonics prediction tasks as training objectives for the pre-training and fine-tuning. In the experiment, each model was fine-tuned on the same dataset\footnote{https://github.com/gitabtion/SoftMaskedBert-PyTorch}.
    \item ChatGPT: We instructed ChatGPT to comprehend sentences holistically and perform Chinese spelling correction, outputting only the corrected sentence without any explanation. To minimize the bias introduced by different prompts, we conducted tests with three sets of zero-shot and few-shot prompts. Detailed prompt settings are in Appendix~\ref{app:Prompt settings}. The output results are obtained using the ChatGPT API. To ensure the stability of the output each time, the temperature is set to 0.
\end{enumerate}

% for 10 epochs
% 为了对比评估LLMs的性能，我们选择了三个强大的CSC模型作为我们的基线：BERT是双向语言模型，其使用分类器为每个字符从词表中选出修正字符输出。SoftMaskedBERT解决BERT没有足够的能力检测每个位置是否存在错误的问题，它包括一个错误检测网络和一个基于BERT的错误校正网络。PLOME将拼音和笔画作为预训练语言模型以及模型微调的输入，设计字符预测任务和拼音预测任务作为预训练语言模型以及模型微调的训练目标。实验中每个模型均在相同数据上微调10个epoch，然后在测试集上测试。

% Table~\ref{tab:Zero-shot and few-shot prompt} shows the settings for the few-shot and zero-shot prompts.
% 我们分别从zero-shot和few-shot两种角度测试ChatGPT在CSC任务中的能力，表2展示了few-shot和zero-shot 模版的设定。具体来说，我们告诉ChatGPT系统从整体上理解句子并完成中文拼写纠错工作，只输出修正后的版本，不做任何解释。同时为了避免提示词注入的问题，使用特殊符号将输入和输出句子间隔开。为更好地对比和观测ChatGPT的纠错性能、降低不同提示词带来的纠错性能的偏差，我们各测试了5组不同的zero-shot和few-shot提示词，其中每一组zero-shot和few-shot的instruction是一致的，few-shot提示词中的example是从训练集中随机采样的，具体提示词设定见附录a。输出结果利用ChatGPT API得到，为保证每次输出结果的稳定性，温度设置为0.

\textbf{Metrics} We computed Eval-CSC based on Script \cite{https://doi.org/10.48550/arxiv.2211.08788}, and implemented Eval-GCSC following the procedures outlined in Section~\ref{sec:Calculation of Eval-GCSC}. Considering that the performance in error correction, rather than error detection, better reflects the proficiency of text spelling correction in CSC, we only retained the metrics results under the correction sub-task. Semantic similarity was evaluated using the three approaches described in Section~\ref{sec:Semantic similarity}.
% Notably, unlike work \cite{PLOME}, our correction metrics are not predicated on detection but are derived from a comprehensive comparison of the source sentence, the reference sentence, and the predicted sentence. 
% 我们按照小节a实现Eval-GCSC，并使用脚本a计算Eval-CSC。考虑到在CSC任务中纠错性能相比检错性能更能表现出文本拼写纠错能力的高低，故只保留了纠错子任务下的指标结果。值得注意的是，不同于工作a，我们计算的纠错指标不是建立在检错基础上的，而是综合对比原句、真值句子和预测句子得到纠错结果。

\subsection{Main Results}\label{sec:new metric result}

\textbf{Comparison of Metric Performance} Table~\ref{tab:Eval-CSC VS Eval-GCSC} displays the comparative evaluation results of ChatGPT under Eval-CSC and Eval-GCSC. Our metric, Eval-GCSC, numerically aligns more closely with human evaluation results than Eval-CSC. The issue is that Eval-CSC can only compute equal-length substitutions, and only considers a prediction correct if it perfectly matches the reference character. This results in correct unequal-length substitutions and reasonable modifications that do not match the reference character being overlooked by Eval-CSC. However, these are retained and appropriately evaluated in both Eval-GCSC and manual assessments.
% 原因在于Eval-CSC只能计算等长替换，且只有预测字符与目标字符完全一致才算正确，这导致正确的非等长替换，以及与目标字符不一致但合理的修改会被Eval-CSC遗漏，但在Eval-GCSC以及人工评估中则均会保留并合理评估。

\begin{figure}
    \centering
    \includegraphics[width=1\linewidth]{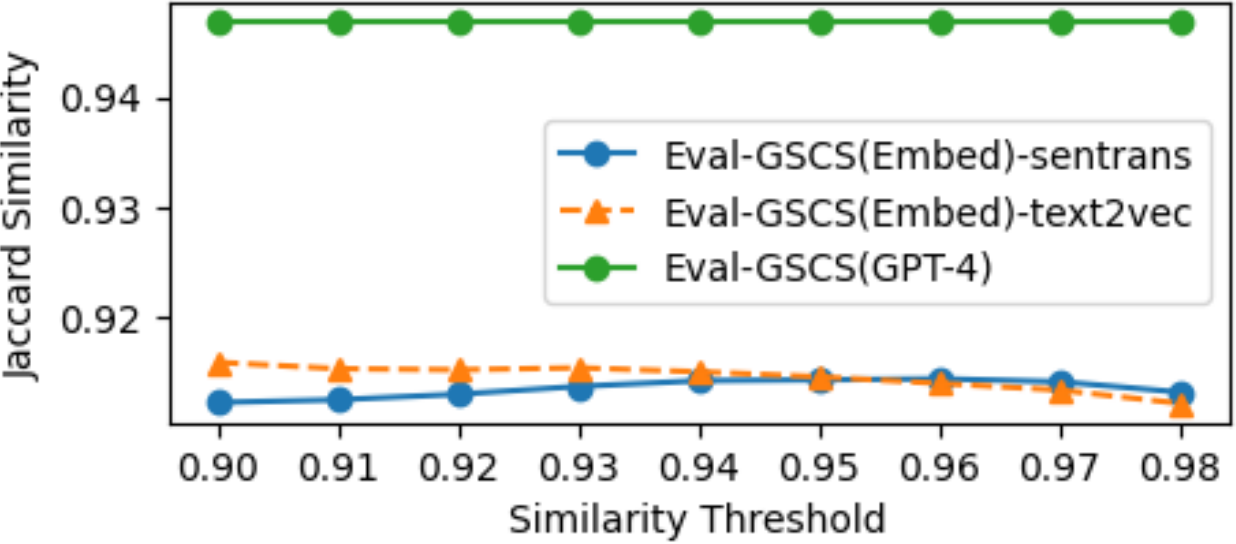}
    \caption{\label{fig:Jaccard similarity}The Jaccard similarity coefficient of Eval-GCSC at different similarity thresholds.}
\end{figure}

To further validate the effectiveness of our metric, we use Eval-GCSC(Embed) and Eval-GCSC(GPT-4) to evaluate the correction results of ChatGPT, and calculate the correlation with human evaluation (Eval-GCSC(Human)). We choose the Jaccard similarity coefficient to measure the correlation. Figure~\ref{fig:Jaccard similarity} shows the Jaccard similarity coefficient of Eval-GCSC at different similarity thresholds. The results show that Eval-GCSC(GPT-4) correlates highly with human evaluation. Eval-GCSC(Embed) using Sentence-Transformer and Text2Vec sentence vectors reach maximum correlation at similarity thresholds of 0.96 and 0.9. This indicates that our metric closely approximates human evaluation, and the introduction of GPT-4 can enhance the performance.

% Eval-GCSC与人工评价在不同相似性阈值下的Jaccard相似系数

\textbf{Comparison of ChatGPT and Baseline} Table~\ref{tab:Eval-CSC VS Eval-GCSC} presents the evaluation results of various models on Eval-CSC and Eval-GCSC. Compared to the results under Eval-CSC, both ChatGPT and the baseline show significant improvements in error correction performance under Eval-GCSC. It indicates that the baseline models' predictions, which include correct non-homophonic length-equivalent substitutions, are appropriately evaluated under Eval-GCSC. This phenomenon suggests that Eval-GCSC can also effectively assess traditional token-level classification models.

% 这表明baseline这些模型得到的预测句中，存在 正确的非音近等长替换 在Eval-GCSC下被合理评估，这个现象说明Eval-GCSC也可以兼顾评估字符级分类模型。

Moreover, under Eval-GCSC, ChatGPT exhibits a more significant enhancement, which makes the error correction ability of ChatGPT not much different from that of token-level classification models. This observation aligns with the conclusions drawn from the human evaluations in \cite{csc_chatgpt}. These findings suggest that Eval-GCSC, compared to Eval-CSC, can more accurately and objectively reflect ChatGPT's correction ability. From another perspective, ChatGPT demonstrates comparable correction performance to the fine-tuned traditional character-level classification models, even without fine-tuning and relying solely on zero-shot and few-shot prompts. This indicates that ChatGPT indeed holds potential as a promising CSC tool.
% <Eval-CSC和Eval-GCSC结果对比> 表a展示了各个模型分别在传统指标Eval-CSC和我们的指标Eval-GCSC结下的评价结果。相比Eval-CSC下的结果，在Eval-GCSC下ChatGPT和baseline的纠错性能均有明显的提升。此外，ChatGPT的提升更多，这使得在我们的评估指标Eval-GCSC下，ChatGPT的纠错能力与传统微调模型相差并不大，这个现象与论文a中人工评价的结论一致。这个结果一方面说明我们的指标相比传统指标能更真实客观地反映ChatGPT的校正能力。

% 另一个角度，ChatGPT在没有微调，只通过zero-shot和few-shot提示就得到与微调后的传统的字符级分类模型可比的性能，这说明ChatGPT确实是一个有潜力的CSC工具。

\begin{table*}[htbp]
    \centering
    \resizebox{\linewidth}{!}{
    \begin{tabular}{c|ccc|ccc|ccc}
    \hline
    \multirow{2}{*}{Model} & \multicolumn{3}{c|}{Eval-CSC} & \multicolumn{3}{c|}{Eval-GCSC (GPT-4)}  & \multicolumn{3}{c}{Eval-GCSC (human)} \\ \cline{2-10} 
                        & P      & R      & F1     & P      & R      & F1   & P      & R      & F1 \\ \hline
    BERT (finetuning)                & 81.085  & 82.460  & 81.767  & 94.785  & 85.574  & 89.944  & 90.728  & 83.893  & 87.177  \\
    SMBERT (finetuning)      & 83.535  & 84.391  & 83.961  & 95.788  & 86.587  & 90.955  & 93.221  & 85.275  & 89.071 \\
    PLOME (finetuning)               & 83.770  & 84.430  & 84.099  & 95.405  & 86.248  & 90.596 & / & / & /  \\ \hline
    ChatGPT (few-shot)  & 60.990  & 64.301  & 62.580  & 94.037  & 76.091  & 84.101 & 89.044  & 72.279  & 79.790   \\ 
    ChatGPT (zero-shot) & 51.450  & 62.173  & 56.224  & 93.926  & 74.453  & 83.052 & 86.538  & 70.796  & 77.879   \\ \hline
    \end{tabular}
    }
    \caption{\label{tab:Eval-CSC VS Eval-GCSC}The error correction performance of various models under Eval-CSC and Eval-GCSC. Eval-GCSC (GPT-4) denotes the use of GPT-4 for semantic similarity assessment. This table only presents the results of the character-level error correction subtasks under Eval-CSC and Eval-GCSC (GPT-4). The results of ChatGPT are the average evaluation value under multiple prompts. Detailed results can be found in Appendix~\ref{app:Detail result of Eval-CSC and Eval-GCSC}.}
\end{table*}
% 各个模型在Eval-CSC和Eval-GCSC下的纠错性能对比。Eval-GCSC(GPT-4)表示使用GPT-4进行语义相似性评估。本表格只展示了Eval-CSC和Eval-GCSC(GPT-4)字符级纠错子任务的结果。表格中的数据是ChatGPT在多个few-shot提示词下的平均评估结果。使用句向量相似性评估的Eval-GCSC(Embed)等详细结果见附录a。

\subsection{Analysis}

\textbf{Rationality of Eval-GCSC} To elucidate why Eval-GCSC is more suitable than Eval-CSC for assessing the performance of generative models in CSC task, we compared the evaluation results of ChatGPT under zero-shot prompt using both metrics. In the following, we analyze from two perspectives: phonics and length.
% 下面我们分别从拼音和长度两个角度分析。其中在Eval-CSC下判断有误的等长操作大概率是非音近的。
\begin{enumerate}[(1)]
    \item Phonetic Perspective: Since Eval-CSC can only evaluate under length constraints, we selected equal-length replacement operations from all correction operations for comparison. Under Eval-CSC, erroneous equal-length operations are likely to be non-homophonic. Table~\ref{tab:case-study-metric-stat} shows the number of operations under various conditions. It indicates that Eval-CSC cannot accurately evaluate corrections that may be correct but do not conform to the phonics constraints, leading to misjudgments of approximately 41.2\% equal-length operations. In contrast, Eval-GCSC can effectively evaluate whether or not they meet the phonics constraints.
    \item Length Perspective: Unequal-length operations are entirely ignored under Eval-CSC, but are fully retained under Eval-GCSC. According to statistics, about 95.8\% of unequal-length correction operations were evaluated as correct by Eval-GCSC (GPT4). This suggests that unequal-length corrections (including character additions, deletions, and unequal-length replacements) are generally acceptable as long as they do not affect the sentence's semantics, which aligns with our intuition.
\end{enumerate}
In summary, Eval-GCSC can more reasonably assess ChatGPT's error correction capability for both non-homophonic and unequal-length operations, compared to Eval-CSC.

% <Eval-GCSC的合理性> 为了解释为什么我们的指标Eval-GCSC比原指标Eval-CSC更适合评估生成式模型在CSC任务下的效果，我们对比了ChatGPT在zero-shot下的两种指标的评价结果。由于Eval-CSC只能在长度约束下评估，故只选取所有纠错操作中的等长替换操作对比，表a展示了各种情况下的操作数。结果显示Eval-CSC对不符合拼音约束但可能正确的纠正操作无法正确评价，造成对大约1200个等长操作的误判，而无论是否满足拼音约束，Eval-GCSC均能很好的评估。
% 非等长的操作在Eval-CSC下会被全部忽略，但在Eval-GCSC下会被全部保留。经统计，有大约95.8%的非等长纠正操作被Eval-GCSC(GPT4)评估为正确。这说明非等长纠正（包含增加字符、删除字符和非等长替换），在不影响句子语义情况下大多是可以接受的，这与我们的直觉接近。
% 总之，Eval-GCSC无论对非音近替换还是非等长替换，均能比Eval-CSC更合理地评估ChatGPT的纠错能力。

% 相同模型下两种指标对
\begin{table}[!ht]
\begin{center}
\resizebox{\linewidth}{!}{
\begin{tabular}{c|c|c}
\hline
Eval-GCSC(GPT4) & Eval-CSC & Percentage of Operations \\ \hline
Correct         & Correct  & 54.1\%              \\ 
Wrong           & Wrong    & 4.7\%               \\ \hline
Correct         & Wrong    & 41.2\%              \\ 
Wrong           & Correct  & 0                 \\ \hline
\end{tabular}
}
\caption{\label{tab:case-study-metric-stat}The percentage of operations corresponding to different evaluation results under ChatGPT.}
\end{center}
\end{table}
% The model chosen is ChatGPT under zero-shot prompt.
% 同一模型下两种指标不同评估结果对应的操作数。模型选择的是zero-shot下的ChatGPT。
% 经统计，Eval-GCSC和Eval-CSC均判断为正确的纠正操作有1574个，Eval-GCSC和Eval-CSC均判断为有误的纠正操作有138个，Eval-GCSC判断正确而Eval-CSC判断有误的有1200个，不存在Eval-CSC判断正确而Eval-GCSC判断有误的情况。
% {'all_right': 1574, 'all_wrong': 138, 'old_right_new_wrong': 0, 'new_right_old_wrong': 1200} 2912

% \textbf{Analysis of ChatGPT's Correction Capability} 
Furthermore, to analyze why ChatGPT's performance in the CSC task under Eval-GCSC evaluation is close to that of token-level classification models, we conducted a detailed observation and comparison of the error correction results of the two models under the same evaluation metric, Eval-GCSC (GPT-4). To better compare ChatGPT and the token-level classification models, we selected and analyzed all the corrective operations for which both models modify the same position of the source sentence. Detailed case studies can be found in the Appendix~\ref{app:case study}. And the summary of the advantages and disadvantages of ChatGPT is as follows:

% 对ChatGPT优缺点的总结如下。更多的Case study见附录。
\begin{CJK*}{UTF8}{gbsn}
\begin{enumerate}[(1)]
    % \begin{itemize}[itemsep=2pt]
    \item Advantages: In the scenario where ChatGPT corrects correctly, and the classification model corrects incorrectly, we found that in 60\% of the cases, ChatGPT's corrections were semantically more appropriate than classification models. For example, "他\textcolor{red}{题}(\textcolor{blue}{提})很多问题" is corrected as "他\textcolor{blue}{提了}很多问题" by ChatGPT, while corrected as "他\textcolor{red}{有}很多问题" by classification model. In another 20\% of the cases, the reference sentences were semantically ambiguous, but ChatGPT made more reasonable modifications. For example, ChatGPT correct "以后我\textcolor{blue}{回来}我的国家" as "以后我\textcolor{blue}{回}我的国家". The remaining few cases demonstrated that ChatGPT could better modify two consecutive typos (correct "脸\textcolor{red}{百百}的" as "脸\textcolor{blue}{白白}的") and correct common knowledge errors more effectively (For exmple, correct "水\textcolor{red}{虑}(\textcolor{blue}{滤})可机" as "水\textcolor{blue}{过滤}机").
    \item Disadvantages: In the scenario where ChatGPT corrects incorrectly and the classification model corrects correctly, We found that in 61\% of the cases, ChatGPT tended to change the original meaning of the sentence, sometimes creating words based on the first character of the incorrect word (correct "我\textcolor{red}{以}(\textcolor{blue}{一})位邻居" as "我\textcolor{red}{以为}邻居"), and sometimes modifying the semantics based on its own understanding (correct "你是怎\textcolor{red}{码}(\textcolor{blue}{么})样" as "你是\textcolor{red}{什么}样"). In cases where the semantics were not changed, ChatGPT made incorrect corrections (accounting for 26\%). In a few cases, ChatGPT made the semantics inappropriate due to the addition or deletion of characters (For example, correct "\textcolor{red}{刻}(\textcolor{blue}{克})服" as "\textcolor{red}{刻苦}服").
    \item Other: In the scenario where both ChatGPT and the classification model correct incorrectly, we found that these cases mainly involved source sentences with the following issues: uncommon knowledge errors, uncommon words, two consecutive typos causing unclear word meanings, unclear source sentence semantics, and typos that could have multiple reasonable corrections. For example, The name of people in "他是杨赵宁" is uncommon knowledge. And the semantics of "我还希望有说中文很里的机会" is itself unclear.
    % \end{itemize}
\end{enumerate}
\end{CJK*}

In summary, the robust language processing capabilities of ChatGPT provide it with certain advantages in performing CSC tasks. However, its limitations may stem from its optimization strategy, which leans more towards generating fluent and natural text rather than ensuring strict spelling accuracy. Additionally, its unidirectional generation method, which does not allow for backtracking and modification after text generation, may limit its error correction capabilities.
% <ChatGPT是有潜力的CSC工具> 为了分析Eval-GCSC评估下ChatGPT在CSC任务中与微调小模型的性能接近的原因，我们对同一评价指标Eval-GCSC(GPT-4)下ChatGPT和微调模型的纠错结果进行了详细的观察对比。为更好地对比ChatGPT和微调模型，我们选取所有两种模型对原句相同位置进行修改的纠正操作，并分下面三种情况分析：
% 1》ChatGPT纠正正确，微调模型纠正错误：我们发现60%的样例中chatgpt的纠正操作比小模型语义更合适，另外20%的样例的参考句语义模糊，但chatgpt改得更合理；其余少部分样例则体现chatgpt对连续两个错别字可以更好地修改，且chatgpt对知识性的错误能更好地纠错。
% 2〉ChatGPT纠正错误，微调模型纠正正确：61%的情况下chatgpt倾向改变句子原意，有时会顺着错误词的第一个字造词，有时则按照自己的理解修改语义；不改变语义情况下，chatgpt存在纠正不正确的情况（占26%）；还有少部分样例中，chatgpt会由于增加或删除字符导致语义不合适。
% 3》ChatGPT和微调模型均纠正错误：这些样例主要是原句存在如下情况：知识型错误、不常用的词语、连着两个错别字造成词语意思不清晰、原句语义不清晰、错别字位置可能存在多种合理的修改。

% 总之，ChatGPT极强的语言处理能力使其做CSC任务具有一定的优势，但仍存在不足的原因可能是其优化策略更偏向于生成流畅、自然的文本，而不是严格的拼写准确性，另外其单向生成的方式，无法在生成文本后进行回溯和修改，这也可能限制了其纠错能力。

\section{Conclusion}
In this work, we believe that the discrepancy between traditional metrics and human evaluation in assessing the error correction performance of ChatGPT is due to the inadequacy of traditional metrics for evaluating generative correction models. Upon analysis, we found that the traditional metrics are flawed, in part, due to the strong constraints on length and phonics, which tend to underestimate ChatGPT's performance. Moreover, these constraints do not necessarily correlate with the aspects of concern (the operation content and correctness) in human evaluation. 

To more reasonably and comprehensively evaluate generative models in the CSC tasks, we propose a new CSC evaluation metric, Eval-GCSC, modeled after the human evaluation process. By extracting word-level operations and assessing semantic similarity, Eval-GCSC relaxes the constraints on length and phonics, preserving all of ChatGPT's correction operations while considering the semantic appropriateness of each operation. Experimental results show that our metric closely aligns with human evaluation, making them suitable for generative error correction models as well as traditional token-level classification models. Furthermore, under this new metric, ChatGPT's error correction performance is comparable to that of token-level classification models, suggesting that ChatGPT is a promising tool for CSC.

% 在本工作中，我们认为造成传统指标和人工评价对ChatGPT纠错性能评估不一致的原因是传统评价指标并不适合评估生成式纠错模型。经过分析发现传统指标不合理的原因，一方面是由于长度和拼音的强约束会造成对ChatGPT性能的低估，另外这种强约束跟人工评估所关心的（操作内容和正确性）没有必然联系。为更合理更全面地评估CSC任务中的生成式模型，仿照人工评价的过程，我们针对生成模型提出了一个新的CSC评估指标Eval-GCSC。其通过词级别操作提取和语义相似性评估放宽了长度和拼音约束，保留ChatGPT所有纠正操作同时考虑每个操作语义的合理性。实验表明，我们的指标接近人工评价，适用于生成式纠错模型的同时也能兼容传统的模型。此外，在新指标下ChatGPT纠错性能与分类模型可比，说明ChatGPT是一个有潜力的CSC工具。

\bibliography{anthology,custom}

\begin{thebibliography}{23}
\expandafter\ifx\csname natexlab\endcsname\relax\def\natexlab#1{#1}\fi

\bibitem[{Cheng et~al.(2020)Cheng, Xu, Chen, Jiang, Wang, Wang, Chu, and Qi}]{cheng-etal-2020-spellgcn}
Xingyi Cheng, Weidi Xu, Kunlong Chen, Shaohua Jiang, Feng Wang, Taifeng Wang, Wei Chu, and Yuan Qi. 2020.
\newblock \href {https://doi.org/10.18653/v1/2020.acl-main.81} {{S}pell{GCN}: Incorporating phonological and visual similarities into language models for {C}hinese spelling check}.
\newblock In \emph{Proceedings of the 58th Annual Meeting of the Association for Computational Linguistics}, pages 871--881, Online. Association for Computational Linguistics.

\bibitem[{Devlin et~al.(2019)Devlin, Chang, Lee, and Toutanova}]{devlin-etal-2019-bert}
Jacob Devlin, Ming-Wei Chang, Kenton Lee, and Kristina Toutanova. 2019.
\newblock \href {https://doi.org/10.18653/v1/N19-1423} {{BERT}: Pre-training of deep bidirectional transformers for language understanding}.
\newblock In \emph{Proceedings of the 2019 Conference of the North {A}merican Chapter of the Association for Computational Linguistics: Human Language Technologies, Volume 1 (Long and Short Papers)}, pages 4171--4186, Minneapolis, Minnesota. Association for Computational Linguistics.

\bibitem[{Fang et~al.(2023)Fang, Yang, Lan, Wong, Hu, Chao, and Zhang}]{Grammar_chatgpt1}
Tao Fang, Shu Yang, Kaixin Lan, Derek~F. Wong, Jinpeng Hu, Lidia~S. Chao, and Yue Zhang. 2023.
\newblock Is chatgpt a highly fluent grammatical error correction system? {A} comprehensive evaluation.
\newblock \emph{CoRR}, abs/2304.01746.

\bibitem[{He and Garner(2023)}]{he2023can}
Mutian He and Philip~N Garner. 2023.
\newblock Can chatgpt detect intent? evaluating large language models for spoken language understanding.
\newblock \emph{arXiv preprint arXiv:2305.13512}.

\bibitem[{Hong et~al.(2019)Hong, Yu, He, Liu, and Liu}]{hong-etal-2019-faspell}
Yuzhong Hong, Xianguo Yu, Neng He, Nan Liu, and Junhui Liu. 2019.
\newblock \href {https://doi.org/10.18653/v1/D19-5522} {{FASP}ell: A fast, adaptable, simple, powerful {C}hinese spell checker based on {DAE}-decoder paradigm}.
\newblock In \emph{Proceedings of the 5th Workshop on Noisy User-generated Text (W-NUT 2019)}, pages 160--169, Hong Kong, China. Association for Computational Linguistics.

\bibitem[{Hu et~al.(2022)Hu, Meng, and Zhou}]{https://doi.org/10.48550/arxiv.2211.08788}
Yong Hu, Fandong Meng, and Jie Zhou. 2022.
\newblock \href {https://doi.org/10.48550/ARXIV.2211.08788} {Cscd-ime: Correcting spelling errors generated by pinyin ime}.

\bibitem[{Jiao et~al.(2018)Jiao, Sun, and Sun}]{jiao2018LAC}
Zhenyu Jiao, Shuqi Sun, and Ke~Sun. 2018.
\newblock \href {https://arxiv.org/abs/1807.01882} {Chinese lexical analysis with deep bi-gru-crf network}.
\newblock \emph{arXiv preprint arXiv:1807.01882}.

\bibitem[{Lewis et~al.(2020)Lewis, Liu, Goyal, Ghazvininejad, Mohamed, Levy, Stoyanov, and Zettlemoyer}]{lewis-etal-2020-bart}
Mike Lewis, Yinhan Liu, Naman Goyal, Marjan Ghazvininejad, Abdelrahman Mohamed, Omer Levy, Veselin Stoyanov, and Luke Zettlemoyer. 2020.
\newblock \href {https://doi.org/10.18653/v1/2020.acl-main.703} {{BART}: Denoising sequence-to-sequence pre-training for natural language generation, translation, and comprehension}.
\newblock In \emph{Proceedings of the 58th Annual Meeting of the Association for Computational Linguistics}, pages 7871--7880, Online. Association for Computational Linguistics.

\bibitem[{Li et~al.(2023)Li, Huang, Ma, Jiang, Li, Zhou, Zheng, and Zhou}]{csc_chatgpt}
Yinghui Li, Haojing Huang, Shirong Ma, Yong Jiang, Yangning Li, Feng Zhou, Hai{-}Tao Zheng, and Qingyu Zhou. 2023.
\newblock On the (in)effectiveness of large language models for chinese text correction.
\newblock \emph{CoRR}, abs/2307.09007.

\bibitem[{Liu et~al.(2021)Liu, Yang, Yue, Zhang, and Wang}]{PLOME}
Shulin Liu, Tao Yang, Tianchi Yue, Feng Zhang, and Di~Wang. 2021.
\newblock {PLOME:} pre-training with misspelled knowledge for chinese spelling correction.
\newblock In \emph{{ACL/IJCNLP} {(1)}}, pages 2991--3000. Association for Computational Linguistics.

\bibitem[{Reimers and Gurevych(2020)}]{reimers-gurevych-2020-making}
Nils Reimers and Iryna Gurevych. 2020.
\newblock \href {https://doi.org/10.18653/v1/2020.emnlp-main.365} {Making monolingual sentence embeddings multilingual using knowledge distillation}.
\newblock In \emph{Proceedings of the 2020 Conference on Empirical Methods in Natural Language Processing (EMNLP)}, pages 4512--4525, Online. Association for Computational Linguistics.

\bibitem[{Tseng et~al.(2015)Tseng, Lee, Chang, and Chen}]{tseng2015introduction}
Yuen-Hsien Tseng, Lung-Hao Lee, Li-Ping Chang, and Hsin-Hsi Chen. 2015.
\newblock Introduction to sighan 2015 bake-off for chinese spelling check.
\newblock In \emph{Proceedings of the Eighth SIGHAN Workshop on Chinese Language Processing}, pages 32--37.

\bibitem[{Wang et~al.(2019)Wang, Tay, and Zhong}]{wang-etal-2019-confusionset}
Dingmin Wang, Yi~Tay, and Li~Zhong. 2019.
\newblock \href {https://doi.org/10.18653/v1/P19-1578} {Confusionset-guided pointer networks for {C}hinese spelling check}.
\newblock In \emph{Proceedings of the 57th Annual Meeting of the Association for Computational Linguistics}, pages 5780--5785, Florence, Italy. Association for Computational Linguistics.

\bibitem[{Wang et~al.(2023)Wang, Liang, Meng, Shi, Li, Xu, Qu, and Zhou}]{wang2023chatgpt}
Jiaan Wang, Yunlong Liang, Fandong Meng, Haoxiang Shi, Zhixu Li, Jinan Xu, Jianfeng Qu, and Jie Zhou. 2023.
\newblock Is chatgpt a good nlg evaluator? a preliminary study.
\newblock \emph{arXiv preprint arXiv:2303.04048}.

\bibitem[{Wu et~al.(2023)Wu, Wang, Wan, Jiao, and Lyu}]{Grammar_chatgpt2}
Haoran Wu, Wenxuan Wang, Yuxuan Wan, Wenxiang Jiao, and Michael~R. Lyu. 2023.
\newblock Chatgpt or grammarly? evaluating chatgpt on grammatical error correction benchmark.
\newblock \emph{CoRR}, abs/2303.13648.

\bibitem[{Wu et~al.(2013)Wu, Liu, and Lee}]{wu2013chinese}
Shih-Hung Wu, Chao-Lin Liu, and Lung-Hao Lee. 2013.
\newblock Chinese spelling check evaluation at sighan bake-off 2013.
\newblock In \emph{Proceedings of the Seventh SIGHAN Workshop on Chinese Language Processing}, pages 35--42.

\bibitem[{Xu et~al.(2021)Xu, Li, Zhou, Li, Wang, Cao, Huang, and Mao}]{xu-etal-2021-read}
Heng-Da Xu, Zhongli Li, Qingyu Zhou, Chao Li, Zizhen Wang, Yunbo Cao, Heyan Huang, and Xian-Ling Mao. 2021.
\newblock \href {https://doi.org/10.18653/v1/2021.findings-acl.64} {Read, listen, and see: Leveraging multimodal information helps {C}hinese spell checking}.
\newblock In \emph{Findings of the Association for Computational Linguistics: ACL-IJCNLP 2021}, pages 716--728, Online. Association for Computational Linguistics.

\bibitem[{Xu(2023)}]{Text2vec}
Ming Xu. 2023.
\newblock Text2vec: Text to vector toolkit.
\newblock \url{https://github.com/shibing624/text2vec}.

\bibitem[{Yu et~al.(2014)Yu, Lee, Tseng, and Chen}]{yu2014overview}
Liang-Chih Yu, Lung-Hao Lee, Yuen-Hsien Tseng, and Hsin-Hsi Chen. 2014.
\newblock Overview of sighan 2014 bake-off for chinese spelling check.
\newblock In \emph{Proceedings of The Third CIPS-SIGHAN Joint Conference on Chinese Language Processing}, pages 126--132.

\bibitem[{Zhang et~al.(2021)Zhang, Pang, Zhang, Wang, He, Sun, Wu, and Wang}]{Phonetic}
Ruiqing Zhang, Chao Pang, Chuanqiang Zhang, Shuohuan Wang, Zhongjun He, Yu~Sun, Hua Wu, and Haifeng Wang. 2021.
\newblock Correcting chinese spelling errors with phonetic pre-training.
\newblock In \emph{{ACL/IJCNLP} (Findings)}, volume {ACL/IJCNLP} 2021 of \emph{Findings of {ACL}}, pages 2250--2261. Association for Computational Linguistics.

\bibitem[{Zhang et~al.(2020{\natexlab{a}})Zhang, Huang, Liu, and Li}]{zhang-etal-2020-spelling}
Shaohua Zhang, Haoran Huang, Jicong Liu, and Hang Li. 2020{\natexlab{a}}.
\newblock \href {https://doi.org/10.18653/v1/2020.acl-main.82} {Spelling error correction with soft-masked {BERT}}.
\newblock In \emph{Proceedings of the 58th Annual Meeting of the Association for Computational Linguistics}, pages 882--890, Online. Association for Computational Linguistics.

\bibitem[{Zhang et~al.(2020{\natexlab{b}})Zhang, Kishore, Wu, Weinberger, and Artzi}]{BERTScore}
Tianyi Zhang, Varsha Kishore, Felix Wu, Kilian~Q. Weinberger, and Yoav Artzi. 2020{\natexlab{b}}.
\newblock Bertscore: Evaluating text generation with {BERT}.
\newblock In \emph{{ICLR}}. OpenReview.net.

\bibitem[{Zhang et~al.(2022)Zhang, Li, Bao, Li, Zhang, Li, Huang, and Zhang}]{zhang-etal-2022-mucgec}
Yue Zhang, Zhenghua Li, Zuyi Bao, Jiacheng Li, Bo~Zhang, Chen Li, Fei Huang, and Min Zhang. 2022.
\newblock \href {https://doi.org/10.18653/v1/2022.naacl-main.227} {{M}u{CGEC}: a multi-reference multi-source evaluation dataset for {C}hinese grammatical error correction}.
\newblock In \emph{Proceedings of the 2022 Conference of the North American Chapter of the Association for Computational Linguistics: Human Language Technologies}, pages 3118--3130, Seattle, United States. Association for Computational Linguistics.

\end{thebibliography}

\appendix
\clearpage

\section{Appendix}
\label{sec:appendix}

\subsection{\label{app:length and pho eval}ChatGPT Basic Competence Tests}
We used Jieba\footnote{https://github.com/fxsjy/jieba} and Snownlp\footnote{https://github.com/isnowfy/snownlp} to segment words in the training set to obtain Chinese words for testing. In the length generation, phonics generation, and phonics discrimination tests, the length of the words ranged from 1 to 10, with 100 words tested for each length. And in the length discrimination test, 1000 word pairs were randomly sampled, with the lengths of the first and second words in each pair not necessarily matching. 
% 我们用jieba和Snownlp在训练集中分词得到用于测试的中文词语。在长度生成、拼音生成和拼音辨别测试中，词语的长度范围从1到10，每个长度的词语取100个进行测试; 在长度辨别测试中，随机采样1000个词对，词对中第一个词和第二个词的长度不一定一致。
\begin{table}[ht]
\begin{center}
\begin{CJK*}{UTF8}{gbsn}
\resizebox{\linewidth}{!}{
\begin{tabularx}{\columnwidth}{p{50pt}|X}
      \hline
      Test & Prompt\\
      \hline
      Length Generation & "\{词语\}"的汉字长度是多少，只输出结果\\ \hline
      Translation & What is the length of the Chinese characters in "\{word\}", only output the result.)\\
      \hline
      Length Discrimination & "\{词语1\}"与"\{词语2\}"两个汉字的长度是否相等，长度相等输出1，长度不等输出0 \\ \hline
      Translation & Do the lengths of the Chinese characters in "\{word1\}" and "\{word2\}" equal? If equal, output 1, if not equal, output 0.)\\
      \hline
      Phonics Generation & 拼音相近是指拼音编辑距离小于等于1，写出1个和"\{词语\}"这个汉字拼音相近的汉字，只输出结果\\ \hline
      Translation & Phonetic similarity refers to a phonetic edit distance of less than or equal to 1. Write out one Chinese character that is phonetically similar to "\{word\}", only output the result.)\\
      \hline
      Phonics Discrimination & "\{词语1\}"和"\{词语2\}"是否拼音相同，相同输出1，不同输出0\\ \hline
      Translation & Do "\{word1\}" and "\{word2\}" have the same pinyin? If the same, output 1, if different, output 0.)\\
      \hline
\end{tabularx}
}
\end{CJK*}
\caption{\label{tab:length prompt}Prompts for the fundamental capability tests of ChatGPT.}
\end{center}
\end{table}

\subsection{Prompt Setting for GPT-4}
\label{app:prompt gpt4}
We use the following prompt for evaluation, and the experiment obtains results by calling the GPT-4 API. To ensure the stability of the output each time, the temperature is set to 0.
% 我们使用如下的prompt进行评估，实验调用GPT-4的API获得结果。

\begin{CJK*}{UTF8}{gbsn}
\begin{table}[!ht]
\begin{center}
\begin{tabularx}{\columnwidth}{X}
      \hline
      PROMPT\\
      \hline
      比较以下两个句子的含义并判断它们是否一致，若一致输出1，若不一致输出0，只输出数字。\\
        \{ \\
        "task": "semantic\_comparison",\\
        "sentences":\\ \quad\quad \{  \\
        \quad\quad\quad "sentence1": "我的名字是张小春", \\
        \quad\quad\quad "sentence2": "我的名字是张晓春" \\
         \quad\quad \}\\
        \}\\
      \hline
      Translation of Instruction\\
      \hline
      Compare the meanings of the following two sentences and judge whether they are consistent. If they are consistent, output 1. If they are inconsistent, output 0. Note that only numbers should be output.\\
      \hline
\end{tabularx}
\caption{\label{tab:gpt4 prompt}Prompt Setting for GPT-4 evaluation.} % of semantic similarity
\end{center}
\end{table}
\end{CJK*}

\subsection{Human Evaluation}
\label{app:human eval}
We invite three annotators to annotate manually and design templates to facilitate human evaluation of the accuracy of each correction made by the model. Before evaluation, annotators undergo training and must pass a test to proceed to the official annotation stage. During the annotation process, the majority consensus of multiple annotations for a single case is considered the final result. In brief, the annotators involved in this assessment are professionals, and their annotations have been reviewed and are highly reliable. Examples of the templates used in our experiments are as shown in Table~\ref{tab:human evaltable}, with the operations derived from Section~\ref{sec:Operations Extraction}. 
% 在评估之前，先对标注者进行标注培训，当标注测试合格方可进入正式标注环节。在进行标注的过程中，对于同一个案例取多数标注结果为最终结果。

% 用人工评估来评估纠错结果是至关重要的。我们设计模版，让人来对模型每个纠错操作的正确性进行评估。完成本评估的标注人员是专业的，标注结果经过检查且可信度高。我们实验所使用的模版举例如下：
\begin{CJK*}{UTF8}{gbsn}
\begin{table}[!ht]
\begin{center}
\begin{tabularx}{\columnwidth}{X}
      \hline
      Human Evaluation\\
      \hline
      原始：我有另外的事，我不\textcolor{green}{可以}去看你。\\
      参考：我有另外的事，我不\textcolor{green}{可以}去看你。\\
      模型修正操作：\\
      1: 可以 -> 能 \\
      正确的修改编号： \\
      错误的修改编号： \\
      \hline
      Translation\\ \hline
      Primal: I have another matter, I \textcolor{green}{can't} visit you. \\
      Reference: I have another matter, I \textcolor{green}{can't} visit you. \\
      Model correction operation: \\
      1: can -> able \\
      Correct revision number:  \\
      Incorrect revision number: \\
      \hline
\end{tabularx}
\caption{\label{tab:human evaltable}The setting of human evaluation prompt.}
\end{center}
\end{table}
\end{CJK*}

\subsection{Details of Metric design}
\label{app:Metric design}
We calculate sentence-level and character-level precision, recall, and F1 scores as follows: 
\begin{enumerate}
    \item Character-level Precision (C\_C\_P): How many words in the predicted words are correctly. \par % 预测正确的词数量 / 预测词的总数
    \item Character-level Recall (C\_C\_R): How many error words in the source sentence have been modified accurately.\par % 预测正确的词 且对应的原词本身包含错别字的词个数 / 含错别字词总数
    \item Sentence-level Precision (S\_C\_P): In all the predicted sentences that have modifications from the source sentence, how many have every single change made correctly, and all errors in the source sentence have been corrected accurately? (If the source sentence does not contain typos, it is not required that all errors in the source sentence be corrected.) \par % 所有对原句有改动的预测句子中，有多少预测句子内的每处改动都正确，且原句所有错误均被修改正确（若原句不含错别字，则不要求原句所有错误均被修改正确）。
    \item Sentence-level Recall (S\_C\_R): Among the source sentences that contain typos, how many have every modification made correctly, and all errors in the source sentence have been accurately corrected.\par % 本身含错别字的原句中，有多少句子中的每处改动都正确，且原句所有错误均被修改正确
    \item Sentence-level F1 (S\_C\_F) or Character-level F1 (C\_C\_F): 2*Precision*Recall / (Precision + Recall).
\end{enumerate}

\subsection{Prompt settings for ChatGPT}
\label{app:Prompt settings}
To avoid the issue of prompt injection, we used special symbols to separate the input and output sentences. To minimize the bias introduced by different prompt words, we conducted tests with three sets of zero-shot and few-shot prompts. In each set, the instructions for both zero-shot and few-shot prompts were consistent. The examples in the few-shot prompts are randomly sampled from the training set, with an equal distribution of five positive and five negative instances. For comparison, we designed 'few\_shot\_0', which uses the same instruction as 'few\_shot\_1'. The difference is that 'few\_shot\_0' has a fixed set of five positive and five negative examples. Detailed instruction settings are in Table~\ref{tab:Zero-shot and few-shot prompt}.

% 为了对比，设计few_shot_0，其与few_shot_1使用相同的instruction，不同的是其中五个正和五个负样例是固定的。

\begin{CJK*}{UTF8}{gbsn}
\begin{table}[htbp]
\begin{center}
\small
\resizebox{\linewidth}{!}{
\begin{tabularx}{\columnwidth}{p{50pt}|X}
      \hline
      ChatGPT & CSC Prompt\\
      \hline
      Instruction(1) \par of zero\_shot\_1 \par few\_shot\_0 \par few\_shot\_1 & 找出并纠正以下标记"input"后的句子中的拼写错误，需要从整体上理解句子，然后在尽可能保持原句子长度不变的情况下逐步发现并纠正拼写错误。直接输出修正后的版本，不做任何解释。\par
      "input":"\{sentence\}"\\
      \hline
      Translation & To find and correct spelling mistakes in the following sentences after the mark "input", you need to understand the sentence as a whole, and then find and correct the spelling mistakes step by step while keeping the length of the source sentence as much as possible. The corrected version will be output directly without any explanation. \par
      "input":"\{sentence\}"\\
      \hline
      Instruction(2) \par of zero\_shot\_2 \par few\_shot\_2 & 仔细阅读原始文本，并找出其中可能存在的错误字符。这些错误可能源于拼音相似性等因素。对于每个错误，用正确的汉字替换，并确保修改后句子长度不变。同时, 尽量使所做的修改符合标准汉字书写规则和惯例，确保关注细节并保持准确度，以实现最佳结果。用标签格式化修正输出结果:"output":你的修正版本。 \par
      "input":"\{sentence\}" =>
      \\ \hline
      Translation & Read the original text carefully and identify any erroneous characters that may be present in it. These errors may stem from factors such as pinyin similarity. For each error, replace the character with the correct one and make sure that the length of the sentence remains the same after the change. At the same time, try to make the changes conform to standard Chinese character writing rules and conventions, ensuring attention to detail and maintaining accuracy for the best results. Format the correction output with the label: "output": your corrected version. \par
      "input":"\{sentence\}" => \\
      \hline
      Instruction(3) \par of zero\_shot\_3 \par few\_shot\_3 & 拼写纠错任务是指你应该使用适当的汉字替换原始文本中存在的任何错误字符，错误的类型可能是拼音相似，注意需要保持纠错前后句子长度不变，请尝试使您的修改最大程度地接近标准汉字书写规则和惯例。输出对反引号内的句子拼写纠错后的结果，不输出标点符号。\par
      "原句":"\{sentence\}" => \\
      \hline
      Translation & The spelling correction task means that you should replace any incorrect characters present in the original text with appropriate Chinese characters, the type of error may be pinyin similarity, note that you need to keep the length of the sentence before and after correction unchanged, please try to make your changes as close as possible to the standard Chinese character writing rules and conventions. Output the result after correcting the spelling of sentences within backquotes, without punctuation marks. \par
      "source sentence":"\{sentence\}" =>  \\
      \hline
\end{tabularx}}
\caption{\label{tab:Zero-shot and few-shot prompt} Zero-shot and few-shot settings for ChatGPT}
\end{center}
\end{table}
\end{CJK*}

\subsection{Detail result of Eval-CSC and Eval-GCSC}
\label{app:Detail result of Eval-CSC and Eval-GCSC}
Table~\ref{tab:old metric appendix}$\sim$Table~\ref{tab:gpt4_sim} present the experimental results of various models under Eval-CSC and Eval-GCSC.
% 表格a到b展示了不同模型分别在Eval-CSC和Eval-GCSC下的实验结果。

\begin{table*}[htbp]
    \centering
    \resizebox{\linewidth}{!}{
    \begin{tabular}{c|ccc|ccc|ccccc}
    \hline
        name & S\_C\_p & S\_C\_r & S\_C\_f1 & C\_C\_p & C\_C\_r & C\_C\_f1 & W & R & M & $S_{ne}$ & $S_{e}$ \\ \hline
        few\_shot\_0 & 56.198 & 53.295 & 54.708 & 61.526 & 62.278 & 61.900 & 15 & 226 & 201 & 113 & 2309 \\ \hline
        few\_shot\_1 & 55.438 & 52.662 & 54.014 & 61.851 & 62.948 & 62.395 & 12 & 379 & 206 & 129 & 2348 \\ 
        %few\_shot\_2 & 57.27 & 54.402 & 55.799 & 64.925 & 65.156 & 65.040 & 10 & 228 & 231 & 141 & 2362 \\ 
        few\_shot\_2 & 52.388 & 53.189 & 52.785 & 59.652 & 66.259 & 62.782 & 20 & 311 & 293 & 182 & 2572 \\ 
        %few\_shot\_4 & 57.417 & 53.664 & 55.477 & 64.936 & 63.579 & 64.25 & 12 & 473 & 333 & 123 & 2289 \\ 
        few\_shot\_3 & 56.702 & 53.295 & 54.946 & 61.468 & 63.697 & 62.563 & 8 & 432 & 281 & 151 & 2378 \\ \hline
        zero\_shot\_1 & 51.033 & 48.181 & 49.566 & 55.676 & 60.505 & 57.99 & 14 & 255 & 321 & 145 & 2444 \\ 
        %zero\_shot\_1 & 45.804 & 48.339 & 47.037 & 50.781 & 66.614 & 57.63 & 30 & 370 & 806 & 204 & 2943 \\ 
        zero\_shot\_2 & 44.921 & 46.389 & 45.643 & 49.613 & 63.106 & 55.552 & 27 & 2311 & 452 & 324 & 2879 \\ 
        %zero\_shot\_3 & 45.404 & 47.127 & 46.249 & 50.46 & 64.88 & 56.769 & 38 & 449 & 1122 & 264 & 2877 \\
        zero\_shot\_3 & 46.642 & 46.863 & 46.752 & 49.062 & 62.909 & 55.129 & 13 & 675 & 450 & 176 & 2877 \\ \hline        
        BERT           & 71.717 &	73.115 &	72.409 &	81.085 &	82.46 &	81.767 &	1 &	1 &	0 &	3 &	2511\\
        SMBERT & 75.526 &	75.646 &	75.586 &	83.535 &	84.391 &	83.961 &	0 &	8 &	0 &	2 &	2486 \\
        PLOME          & 74.553	& 74.75	& 74.651	& 83.77	& 84.43 &	84.099 &	1 &	1	& 0	& 2	& 2424 \\ \hline
    \end{tabular}
    }
\caption{\label{tab:old metric appendix}Evaluation results under Eval-CSC.}
\end{table*}

% Eval-CSC下的评价结果。

\begin{table*}[htbp]
    \centering
    \begin{tabular}{c|ccc|ccc}
    \hline
        name & S\_C\_p & S\_C\_r & S\_C\_f1 & C\_C\_p & C\_C\_r & C\_C\_f1 \\ \hline
        few\_shot\_0 & 81.010 & 70.743 & 75.529 & 95.835 & 75.062 & 84.186 \\ \hline
        few\_shot\_1 & 82.158 & 72.430 & 76.988 & 95.885 & 76.620 & 85.177 \\ 
        %few\_shot\_2 & 83.253 & 73.643 & 78.154 & 95.904 & 77.782 & 85.898 \\ 
        few\_shot\_2 & 83.037 & 76.436 & 79.6 & 95.516 & 80.46 & 87.344 \\ 
        %few\_shot\_4 & 82.188 & 72.219 & 76.882 & 96.635 & 76.169 & 85.190 \\ 
        few\_shot\_3 & 83.294 & 73.432 & 78.053 & 96.351 & 77.349 & 85.811 \\ \hline
        zero\_shot\_1 & 81.818 & 69.478 & 75.145 & 95.538 & 74.252 & 83.561  \\ 
        %zero\_shot\_1 & 83.077 & 76.384 & 79.590 & 96.142 & 81.214 & 88.050 \\ 
        zero\_shot\_2 & 81.590 & 74.750 & 78.020 & 96.301 & 79.869 & 87.319 \\ 
        %zero\_shot\_3 & 75.569 & 69.900 & 72.624 & 91.597 & 80.271 & 85.561 \\ 
        zero\_shot\_3 & 78.835 & 71.798 & 75.152 & 94.206 & 77.317 & 84.93 \\ \hline
        BERT & 88.591 & 85.345 & 86.938 & 97.533 & 86.926 & 91.925 \\ 
        SMBERT & 89.891 & 86.452 & 88.138 & 98.195 & 87.818 & 92.717 \\
        PLOME & 90.263 & 86.241 & 88.206 & 98.550 & 87.521 & 92.709 \\ \hline
    \end{tabular}
\caption{Evaluation results under Eval-GCSC. The similarity is calculated by the Sentence-Transformer \cite{reimers-gurevych-2020-making}, with a similarity threshold set at 0.96.}
\label{tab:embedding_sim}
\end{table*}
%Sentence-Transformer \cite{reimers-gurevych-2020-making} or Text2Vec \cite{Text2vec}
% Eval-GCSC下的评价结果。相似性由sentence-transformer计算得到，相似度阈值取0.96.

\begin{table*}[htbp]
    \centering
    \begin{tabular}{c|ccc|ccc}
    \hline
        name & S\_C\_p & S\_C\_r & S\_C\_f1 & C\_C\_p & C\_C\_r & C\_C\_f1 \\ \hline
        few\_shot\_0 & 84.361 & 73.326 & 78.457 & 98.765 & 76.948 & 86.502  \\ \hline
        few\_shot\_1 & 85.471 & 75.171 & 79.991 & 98.65 & 78.507 & 87.433  \\ 
        %few\_shot\_2 & 87.066 & 76.542 & 81.466 & 98.874 & 79.959 & 88.416  \\ 
        few\_shot\_2 & 87.522 & 79.547 & 83.344 & 99.067 & 82.594 & 90.084  \\ 
        %few\_shot\_4 & 84.899 & 74.275 & 79.232 & 98.899 & 77.646 & 86.993  \\ 
        few\_shot\_3 & 86.729 & 76.015 & 81.019 & 99.033 & 78.95 & 87.858  \\ \hline
        zero\_shot\_1 & 86.054 & 72.588 & 78.749 & 98.909 & 76.261 & 86.121  \\ 
        %zero\_shot\_1 & 85.923 & 79.547 & 82.612 & 98.176 & 83.142 & 90.036  \\ 
        zero\_shot\_2 & 86.952 & 78.809 & 82.680 & 99.206 & 82.206 & 89.909  \\ 
        %zero\_shot\_3 & 86.179 & 79.072 & 82.473 & 98.276 & 82.486 & 89.691  \\ 
        zero\_shot\_3 & 83.790 & 76.015 & 79.713 & 97.614 & 79.820 & 87.825  \\ \hline
        BERT & 90.492 & 86.610 & 88.508 & 99.204 & 87.664 & 93.078 \\ 
        SMBERT & 91.328 & 87.243 & 89.239 & 99.398 & 88.187 & 93.457 \\ 
        PLOME & 91.307 & 86.716 & 88.952 & 99.396 & 87.726 & 93.197 \\ \hline
    \end{tabular}
\caption{Evaluation results under Eval-GCSC. The similarity is calculated by the Text2Vec \cite{Text2vec}, with a similarity threshold set at 0.9.}
\label{tab:embedding_sim_text2vec}
\end{table*}

%Eval-GCSC下的评价结果。相似性由text2vec计算得到，相似度阈值取0.9.

\begin{table*}[htbp]
    \centering
    \begin{tabular}{c|ccc|ccc}
    \hline
        name & S\_C\_p & S\_C\_r & S\_C\_f1 & C\_C\_p & C\_C\_r & C\_C\_f1 \\ \hline
        few\_shot\_0 & 79.602 & 68.846 & 73.834 & 93.929 & 72.765 & 82.003  \\ \hline
        few\_shot\_1 & 80.218 & 70.427 & 75.004 & 93.943 & 74.446 & 83.066  \\ 
        %few\_shot\_2 & 81.612 & 71.956 & 76.480 & 93.959 & 75.975 & 84.015  \\ 
        few\_shot\_2 & 81.128 & 74.170 & 77.493 & 93.47 & 78.695 & 85.449  \\ 
        %few\_shot\_4 & 80.552 & 70.480 & 75.180 & 95.189 & 74.528 & 83.601  \\ 
        few\_shot\_3 & 81.553 & 71.323 & 76.096 & 94.697 & 75.133 & 83.788  \\ \hline
        zero\_shot\_1 & 79.819 & 67.001 & 72.850 & 93.451 & 71.710 & 81.150  \\ 
        %zero\_shot\_1 & 80.500 & 72.905 & 76.514 & 93.827 & 77.933 & 85.145  \\ 
        zero\_shot\_2 & 80.465 & 72.219 & 76.119 & 95.119 & 77.286 & 85.280  \\ 
        %zero\_shot\_3 & 82.033 & 73.748 & 77.670 & 95.454 & 78.220 & 85.982  \\ 
        zero\_shot\_3 & 77.749 & 68.74 & 72.967 & 93.208 & 74.364 & 82.726  \\ \hline
        BERT & 85.958 & 83.342 & 84.630 & 94.785 & 85.574 & 89.944 \\ 
        SMBERT & 88.107 & 84.871 & 86.459 & 95.788 & 86.587 & 90.955  \\
        PLOME & 86.935 & 83.553 & 85.210 & 95.405 & 86.248 & 90.596 \\ \hline
    \end{tabular}
\caption{Evaluation results under Eval-GCSC. The similarity is evaluated by the GPT-4.}
\label{tab:gpt4_sim}
\end{table*}
%Eval-GCSC下的评价结果。相似性由GPT4得到.

\subsection{Case Study}
\label{app:case study}

Table~\ref{tab:case-study-metric} provides a comparative overview of the evaluation results under the same model (ChatGPT) for Eval-CSC and Eval-GCSC. Eval-GCSC not only assesses non-equal length correction operations but also evaluates modification operations inconsistent with reference characters through semantic rationality. Table~\ref{tab:case-study-model}, on the other hand, provides examples comparing the error correction results of ChatGPT and a smaller error correction model under the same metric, Eval-GCSC (GPT-4). ChatGPT exhibits instances of character addition and deletion, but there are examples where it correctly modifies the errors made by the smaller model.
% 表a给出一些例子直观对比同一模型（ChatGPT）下Eval-CSC和Eval-GCSC的评估结果。Eval-GCSC不仅可以评估非等长纠正操作，也能通过语义合理评估与参考字符不一致的修改操作；表b则给出示例，对比相同指标Eval-GCSC(GPT4)下，ChatGPT与纠错小模型的纠错结果。其中ChatGPT存在增加删除字符的现象，但存在例子能改对小模型错误修改的位置。

\begin{table*}[htbp]
\begin{center}
\begin{CJK*}{UTF8}{gbsn}
\resizebox{\linewidth}{!}{
\begin{tabular}{l|c|c}
\hline
Cases & Eval-GCSC(GPT4) & Eval-CSC \\ \hline
\begin{tabular}[c]{@{}l@{}}Src: \textcolor{blue}{和友人}一同吃一顿晚餐，也能拥有莫大的快乐。\\ 
%Tran: Having dinner with companions can also be a great joy.\\
Ref: \textcolor{blue}{和友人}一同吃一顿晚餐，也能拥有莫大的快乐。\\ Tran: Having dinner with \textcolor{blue}{companions} can also be a great joy.\\ Pre: \textcolor{blue}{和朋友}一同吃一顿晚餐，也能拥有莫大的快乐。\\ Tran: Having dinner with \textcolor{blue}{friends} can also be a great joy.\end{tabular} &
  Correct &
  Wrong \\ \hline
\begin{tabular}[c]{@{}l@{}}Src: 没有了它，我们就不能在\textcolor{blue}{闲假时}在上面涂鸦。\\ 
%Tran: Without it, we can't doodle on it during our free time.\\ 
Ref: 没有了它，我们就不能在\textcolor{blue}{闲暇时}在上面涂鸦。\\ Tran: Without it, we wouldn't be able to doodle on it at our \textcolor{blue}{leisure}.\\ Pre: 没有了它，我们就不能在\textcolor{blue}{闲暇时}在上面涂鸦。\\ Tran: Without it, we wouldn't be able to doodle on it at our \textcolor{blue}{leisure}.\end{tabular} &
  Correct &
  Correct \\ \hline
\begin{tabular}[c]{@{}l@{}}Src: 我最感谢的人是我\textcolor{blue}{以位}邻居。\\ 
% Tran: The person I'm most grateful to is my neighbor.\\ 
Ref: 我最感谢的人是我\textcolor{blue}{一位}邻居。\\ Tran: The person I'm most grateful to is \textcolor{blue}{one of} my neighbors.\\ Pre: 我最感谢的人是我\textcolor{blue}{以为}邻居。\\ Tran: The person I'm most grateful to is the neighbor I \textcolor{blue}{thought}.\end{tabular} &
  Wrong &
  Wrong \\ \hline
\begin{tabular}[c]{@{}l@{}}Src: 今年没帮你\textcolor{blue}{庆祖}但在这祖您身体健康要多保重。\\ 
%Tran: I didn't help you this year, but you should take care of your health this year.\\ 
Ref: 今年没帮你\textcolor{blue}{庆祝}但在这祝您身体健康要多保重。\\ Tran: I didn't help you \textcolor{blue}{celebrate} this year but I wish you good health and take care.\\ Pre: 今年没帮你\textcolor{blue}{庆祝，}但在这祝您身体健康要多保重。\\ Tran: I didn't help you \textcolor{blue}{celebrate} this year, but I wish you good health and take care.\end{tabular} &
  Correct &
  / \\ \hline
\end{tabular}
}
\caption{\label{tab:case-study-metric}Cases of evaluation results for Eval-CSC and Eval-GCSC under the same model (ChatGPT).}%We mark the \textcolor{red}{wrong}/\textcolor{blue}{correct} characters.}
\end{CJK*}
\end{center}
\end{table*}
% 同一模型（ChatGPT）下Eval-CSC和Eval-GCSC的评估结果。

\begin{table*}[htbp]
\begin{center}
\begin{CJK*}{UTF8}{gbsn}
\resizebox{\linewidth}{!}{
\begin{tabular}{l|c}
\hline
Cases &
  Eval-GCSC(GPT4) \\ \hline
\begin{tabular}[c]{@{}l@{}}Src: 当你一旦经历和\textcolor{red}{刻服}了这些关卡时...\\ 
%Tran: Once you have experienced and overcome these levels.\\ 
Ref: 当你一旦经历和\textcolor{blue}{克服}了这些关卡时...\\ Tran: Once you have experienced and \textcolor{blue}{conquered} these levels.\\ 
Pre\_ChatGPT: 当你一旦经历和\textcolor{red}{刻苦服}了这些关卡时\\ Tran: Once you have experienced and \textcolor{red}{painstakingly overcome} these levels.\\ 
Pre\_SoftBERT: 当你一旦经历和\textcolor{blue}{克服}了这些关卡时\\
Tran: Once you have experienced and \textcolor{blue}{conquered} these levels.\end{tabular} &
  \begin{tabular}[c]{@{}l@{}}ChatGPT: Wrong\\ SoftBERT: Correct\end{tabular} \\ \hline
\begin{tabular}[c]{@{}l@{}}Src: 没有人来给他们\textcolor{red}{结少}网络的用法。\\ 
%Tran: No one to give them less Internet usage.\\ 
Ref: 没有人来给他们\textcolor{blue}{介绍}网络的用法。\\ 
Tran: No one came to \textcolor{blue}{introduce} them to the use of the network.\\
Pre\_ChatGPT: 没有人来给他们\textcolor{blue}{介绍}网络的用法\\ Tran: No one came to \textcolor{blue}{introduce} them to the use of the network.\\ Pre\_SoftBERT: 没有人来给他们\textcolor{red}{减绍}网络的用法。\\ Tran: No one came to \textcolor{red}{reduce their saws} the use of the network.\end{tabular} &
  \begin{tabular}[c]{@{}l@{}}ChatGPT: Correct\\ SoftBERT: Wrong\end{tabular} \\ \hline
\begin{tabular}[c]{@{}l@{}}Src: 以后就要加油，努力\textcolor{red}{迈近}我的人生目标了。\\
%Tran: I have to work hard in the future and strive to approach my life goals.\\ 
Ref: 以后就要加油，努力\textcolor{blue}{迈进}我的人生目标了。\\ Tran: I have to work hard in the future and \textcolor{blue}{strive to approach} my life goals.\\ 
Pre\_ChatGPT: 以后就要加油，努力\textcolor{blue}{迈向}我的人生目标了。\\ 
Tran: I have to work hard in the future and \textcolor{blue}{strive to move towards} my life goals.\\ 
Pre\_SoftBERT: 以后就要加油，努力\textcolor{blue}{迈进}我的人生目标了。\\ 
Tran: I have to work hard in the future and \textcolor{blue}{strive to approach} my life goals.\end{tabular} &
  \begin{tabular}[c]{@{}l@{}}ChatGPT: Correct\\ SoftBERT: Correct\end{tabular} \\ \hline
\begin{tabular}[c]{@{}l@{}}Src: 而且要准备考试时，更不知道应该要\textcolor{red}{付息}什么才好。\\ 
%Tran: And to prepare for the exam, do not know what should be paid interest.\\ 
Ref: 而且要准备考试时，更不知道应该要\textcolor{blue}{复习}什么才好。\\ Tran: And when preparing for the exam, I don't know what to \textcolor{blue}{review}.\\ Pre\_ChatGPT: 而且要准备考试时，更不知道应该要\textcolor{red}{扶持}什么才好。\\ Tran: And when preparing for the exam, I don't know what to \textcolor{red}{support}.\\ Pre\_SoftBERT: 而且要准备考试时，更不知道应该要\textcolor{red}{付习}什么才好。\\ Tran: And when preparing for the exam, I don't know what you should \textcolor{red}{pay} for.\end{tabular} &
  \begin{tabular}[c]{@{}l@{}}ChatGPT: Wrong\\ SoftBERT: Wrong\end{tabular} \\ \hline
\end{tabular}
}
\caption{\label{tab:case-study-model}Cases of error correction comparison results between ChatGPT and the smaller error correction model under the same metric Eval-GCSC (GPT-4). We mark the \textcolor{red}{wrong}/\textcolor{blue}{correct} characters.}
\end{CJK*}
\end{center}
\end{table*}
% 相同指标Eval-GCSC(GPT4)下，ChatGPT与纠错小模型的纠错对比结果。

% # {'all_right': 1873, 'all_wrong': 20, 'chatgpt_right_soft_wrong': 48, 'soft_right_chatgpt_wrong': 92} 2033

% bert_unique_calculate {'all_right': 1823, 'all_wrong': 20, 'chatgpt_right_soft_wrong': 30, 'soft_right_chatgpt_wrong': 92}
% ceshi_bert4csc_result_calculate {'all_right': 1798, 'all_wrong': 20, 'chatgpt_right_bert_wrong': 48, 'bert_right_chatgpt_wrong': 91}
% Plome_calculate {'all_right': 1755, 'all_wrong': 14, 'chatgpt_right_plome_wrong': 38, 'plome_right_chatgpt_wrong': 92}

\end{document}